\documentclass[10pt,twocolumn,letterpaper]{article}
\usepackage{cvpr}              

\usepackage{times}
\usepackage{epsfig}
\usepackage{graphicx}
\usepackage{amsmath}
\usepackage{amssymb}
\usepackage[dvipsnames]{xcolor}
\usepackage{array}
\usepackage{multirow}
\usepackage{mathtools}
\usepackage{overpic}
\usepackage{stmaryrd}
\usepackage{booktabs}

\definecolor{cvprblue}{rgb}{0.21,0.49,0.74}
\usepackage[pagebackref,breaklinks,colorlinks,citecolor=cvprblue]{hyperref}

\usepackage[capitalize]{cleveref}
\crefname{section}{Sec.}{Secs.}
\Crefname{section}{Section}{Sections}
\crefname{table}{Tab.}{Tabs.}
\Crefname{table}{Table}{Tables}
\crefname{figure}{Fig.}{Figs.}
\Crefname{figure}{Figure}{Figures}
\crefname{equation}{Eq.}{Eqs.}
\Crefname{equation}{Equation}{Equations}
\crefname{appendix}{Appx.}{Appxs.}
\Crefname{Appendix}{Appendix}{Appendices}

\def\paperID{416} 
\def\confName{CVPR}
\def\confYear{2024}

\usepackage{amsmath}

\definecolor{DeepPurple}{RGB}{50, 2, 104}

\def\tight#1{\hspace{1pt}{#1}{\hspace{1pt}}}
\def\medium#1{\hspace{2pt}{#1}{\hspace{2pt}}}

\newcolumntype{P}[1]{>{\centering\arraybackslash}p{#1}}
\newlength{\wdth}

\newcommand{\ptitle}[1]{\noindent\textbf{#1}\hspace{5pt}}

\newcommand{\ours}{{MIRETR}}
\begin{document}

\title{Learning Instance-Aware Correspondences for Robust Multi-Instance Point Cloud Registration in Cluttered Scenes}

\author{
Zhiyuan Yu$^1$\hspace{8pt}Zheng Qin$^2$\hspace{8pt}Lintao Zheng$^1$\hspace{8pt}Kai Xu$^1$\thanks{Corresponding author: Kai Xu (kevin.kai.xu@gmail.com).}\\
$^1$ National University of Defense Technology\\
$^2$ Defense Innovation Institute, Academy of Military Sciences
}
\maketitle

\begin{abstract}
Multi-instance point cloud registration estimates the poses of multiple instances of a model point cloud in a scene point cloud.
Extracting accurate point correspondences is to the center of the problem. Existing approaches usually treat the scene point cloud as a whole, overlooking the separation of instances. Therefore, point features could be easily polluted by other points from the background or different instances, leading to inaccurate correspondences oblivious to separate instances, especially in cluttered scenes.
In this work, we propose \textbf{\ours{}}, \textbf{M}ulti-\textbf{I}nstance \textbf{RE}gistration \textbf{TR}ansformer, a coarse-to-fine approach to the extraction of \emph{instance-aware correspondences}.
At the coarse level, it jointly learns instance-aware superpoint features and predicts per-instance masks.
With instance masks, the influence from outside of the instance being concerned is minimized, such that highly reliable superpoint correspondences can be extracted.
The superpoint correspondences are then extended to instance candidates at the fine level according to the instance masks.
At last, an efficient candidate selection and refinement algorithm is devised to obtain the final registrations.
Extensive experiments on three public benchmarks demonstrate the efficacy of our approach. In particular, \ours{} outperforms the state of the arts by $16.6$ points on F$_1$ score on the challenging ROBI benchmark.
Code and models are available at \url{https://github.com/zhiyuanYU134/MIRETR}.
\end{abstract}
\vspace{-10pt}


\section{Introduction}

\label{sec:intro}

Point cloud registration aims at estimating a rigid transformation that aligns two point clouds. In real-world application scenarios such as robotic bin picking, there is often the requirement of multi-instance registration where the point cloud of a model needs to be registered against multiple instances of the model in the target scene. \emph{Multi-instance point cloud registration} is challenging due to the unknown number of instances and inter-instance occlusions especially in a cluttered scene.


\begin{figure}[t]
\begin{overpic}[width=1.0\linewidth]{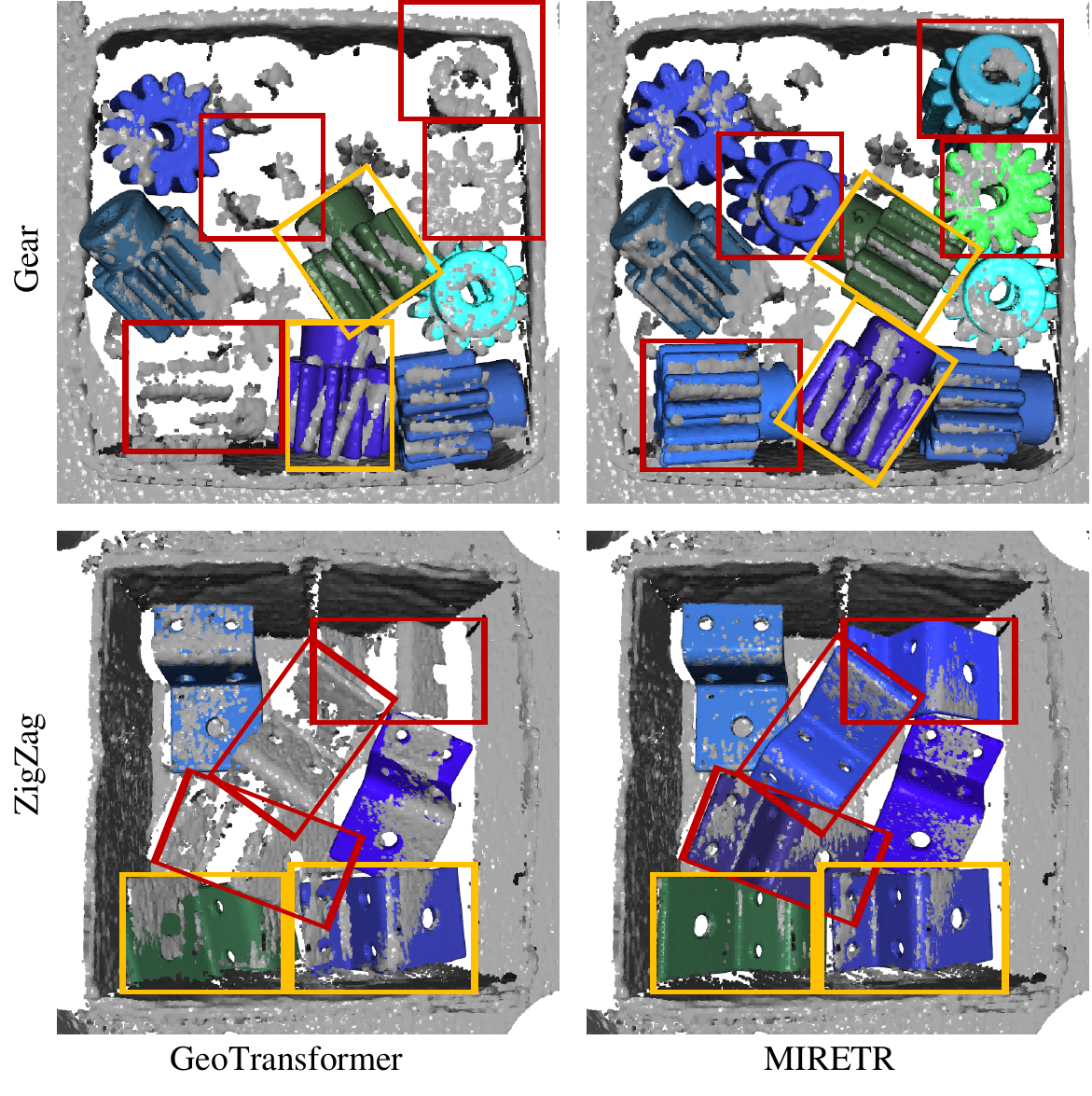}
\end{overpic}
\caption{
\ours{} significantly improves the multi-instance registration results in cluttered scenes compared to the state-of-the-art GeoTransformer~\cite{qin2022geometric}. Benefiting from the instance-aware correspondences, our method can generate more accurate registrations (see the \textcolor{Dandelion}{yellow} boxes) and register the heavily-occluded instances with severe geometric deficiency (see the \textcolor{red}{red} boxes).
}
\label{fig:teaser}
\vspace{-15pt}
\end{figure}

Existing approaches to multi-instance registration are mostly two-stage: They first extract point correspondences between the model and the scene point clouds and then recover per-instance transformations with multi-model fitting~\cite{kluger2020consac,tang2022multi,yuan2022pointclm}. Clearly, the performance of such approach highly depends on the quality of the correspondences.
Unlike pair-wise registration that can be pinned down by a small number of correspondences, multi-instance registration poses unique challenges. First, the correspondences should widely spread over the scene point cloud to cover as many instances as possible so that more instances can be found and registered. Second, the correspondences should be accurately clustered into individual instances to estimate per-instance pose transformations.

Existing point correspondence methods can be broadly classified into \emph{keypoint-based} and \emph{keypoint-free} ones.
Albeit successful in pairwise registration, these methods fail to effectively tackle the aforementioned challenges.
On the one hand, keypoint-based methods~\cite{choy2019fully,bai2020d3feat,huang2021predator} first detect keypoints and then match them to form correspondences. In the multi-instance setting, however, it is common that an instance is severely occluded so that too few keypoints can be detected to guarantee a successful registration.

On the other hand, keypoint-free methods~\cite{yu2021cofinet,qin2022geometric} bypass keypoint detection with a coarse-to-fine pipeline.
They first extract correspondences between sparse superpoints and then refine them to dense point correspondences.
To achieve reliable superpoint matching, they encode global context with feature correlation~\cite{yu2021cofinet} or geometric structure~\cite{qin2022geometric} to learn distinctive superpoint features.
When used in multi-instance registration, such approach may overlook the separation of individual instances and lead to the following two drawbacks.
First, the superpoint features of an instance may be contaminated by the global context from the background or other instances, making the extracted features less informative.
Thus, the features of a severely occluded instance could be overwhelmed by those more complete neighboring ones, making the occluded instances hard to registered (see \cref{fig:teaser}~(left)).
Second, they cannot discriminate correspondences into different instances and have to rely on multi-model fitting~\cite{kluger2020consac,tang2022multi,yuan2022pointclm} to cluster correspondences. Multi-model fitting needs to sample numerous hypotheses and finds difficulty in handling heavy occlusion.

We propose \emph{\textbf{\ours{}}}, \emph{\textbf{M}ulti-\textbf{I}nstance \textbf{RE}gistration \textbf{TR}ansformer}, which learns to extract instance-aware correspondences in a coarse-to-fine fashion.
Our key motivation is to restrict the contextual feature encoding within the instance scope.
An \emph{Instance-aware Geometric Transformer} module is devised to jointly extract reliable \emph{superpoint} features and predict instance masks in the scene.
We resolve this chicken-and-egg problem with an iterative process.
In each iteration, we first encode intra-instance geometric structure based on the instance mask of the previous iteration and then learn geometric consistency between the model and the scene.
Next, the instance masks are refined based on the instance-aware features.
As the iteration goes, the superpoint features gradually improve with increasingly relevant context and accurate instance masks, resulting in reliable registrations even for heavily-occluded instances (see \cref{fig:teaser} (right)).

Having obtained superpoint correspondences, we extend them to instance candidates based on instance masks.
This allows us to directly extract instance-wise \emph{point correspondences} and estimate a transformation for each instance candidate.
An efficient candidate selection and refinement method is proposed to eliminate duplicated instances and obtain the final registrations.
Thanks to the instance-aware correspondences, our method bypasses the need of multi-model fitting, achieving superior efficiency and accuracy.
Extensive experiments on three benchmarks~\cite{avetisyan2019scan2cad,yang2021robi,chang2015shapenet} demonstrate the efficacy of our method.
Our method surpasses the previous state-of-the-art methods~\cite{qin2022geometric,yuan2022pointclm} by $16.6$ points on F$_1$ score on the cluttered ROBI benchmark.
Our main contributions include:
\begin{itemize}
  \item A multi-instance point cloud registration method which extracts instance-wise correspondences and estimates transformations without multi-model fitting.
  \item An instance-aware geometric transformer module which jointly learns instance-aware superpoint features and predicts instance masks.
  \item An efficient instance candidate selection and refinement method which removes duplicated candidates and generates the final registrations.
\end{itemize}


\section{Related work}
\label{sec:related}
\ptitle{Point cloud registration.}
State-of-the-art point cloud registration methods can be categorized into direct registration methods and correspondence-based methods. 
Direct registration methods~\cite{wang2019deep,wang2019prnet,yew2020rpm,fu2021robust,aoki2019pointnetlk,huang2020feature,xu2021omnet} use a neural network to estimate the transformation in an end-to-end manner.
Correspondence-based methods~\cite{deng2018ppfnet,deng2018ppf,gojcic2019perfect,choy2019fully,bai2020d3feat,huang2021predator,yu2021cofinet,qin2022geometric} extract point correspondences and estimate the rigid transformation with a robust pose estimator.
Early point correspondence methods~\cite{bai2020d3feat,huang2021predator} adopt a detect-then-match pipeline, where keypoints are first detected and then matched.
Recent advances~\cite{yu2021cofinet,qin2022geometric} have bypassed keypoint detection by extracting correspondences in a coarse-to-fine fashion.
However, existing methods mainly focus on pairwise registration and little attention has been paid to multi-instance registration, which faces new challenges such as unknown number of instances and heavy intra-instance occlusion.
In this paper, we fill this gap by directly extracting instance-aware correspondences.

\ptitle{Multi-model fitting.}
Multi-model fitting aims at recovering multiple models from noisy data, which is an important step in multi-instance registration.
Multi-model fitting methods can be classified into RANSAC-based methods and clustering-based methods.
RANSAC-based methods~\cite{kanazawa2004detection,barath2019progressive,kluger2020consac,barath2021progressive} adopt a hypothesize-and-verify manner to sequentially fit multiple models.
Clustering-based methods~\cite{magri2014t,magri2015robust,magri2016multiple,tang2022multi,yuan2022pointclm} sample a huge set of hypotheses and cluster the correspondences according to their residuals under these hypotheses.
However, these methods suffer from huge memory usage or long convergence time. In this work, we achieve accurate and efficient multi-instance registration without multi-model fitting.

\ptitle{Open-set pose estimation.}
Model retrieval~\cite{besl1992method,binford1982survey,chin1986model,avetisyan2019scan2cad,avetisyan2020scenecad,huang2018holistic,kuo2020mask2cad,gumeli2022roca} and image-based pose estimation~\cite{xiang2017posecnn,sundermeyer2018implicit,hodan2018bop,hu2020single,shi2021stablepose,ponimatkin2022focal,su2022zebrapose,chen2022sim,wang2019normalized,chen2020learning,lin2021dualposenet,lin2022category} are two highly-related tasks to multi-instance registration.
Model retrieval aims at estimating the $9$-DoF pose parameters of the CAD models from a pre-built database that appears in a given scene.
And image-based pose estimation recovers the $6$-DoF poses for certain objects~\cite{xiang2017posecnn,sundermeyer2018implicit,hodan2018bop,hu2020single,shi2021stablepose,ponimatkin2022focal,su2022zebrapose,chen2022sim} or categories~\cite{wang2019normalized,chen2020learning,lin2021dualposenet,lin2022category} in an RGB or RGB-D image.
However, both tasks are based on the \emph{close-set} assumption, where the training and testing data follow the same distribution. Model retrieval can only retrieve the CAD models in the database while image-based pose estimation can only handle known objects or categories in the training data.
On the contrary, multi-instance registration is an \emph{open-set} problem. It is based on geometric matching and thus can generalize well to novel objects and categories. We discuss the generality of \ours{} in \cref{sec:exp-ShapeNet}.


\section{Problem Statement}
\label{sec:problem-stat}

Given a model point cloud $\mathcal{P} \tight{=} \{ \mathbf{p}_i \tight{\in} \mathbb{R}^3 \tight{\mid} i \tight{=} 1, ..., N\}$ and a scene point cloud $\mathcal{Q} \tight{=} \{\mathbf{q}_i \tight{\in} \mathbb{R}^3 \tight{\mid} i \tight{=} 1, ..., M\}$, where $\mathcal{Q}$ could contain multiple instances of $\mathcal{P}$, the goal of multi-instance registration is to register all instances of $\mathcal{P}$ in $\mathcal{Q}$. Specifically, $\mathcal{Q}$ is represented as $\mathcal{Q} = \mathcal{Q}_0 \cup \mathcal{Q}_1 \cup ... \cup \mathcal{Q}_K$, where $\mathcal{Q}_0$ is the background points and $\{\mathcal{Q}_1, ..., \mathcal{Q}_K\}$ are the $K$ instances of $\mathcal{P}$. For each instance $\mathcal{Q}_k$, we estimate a $6$-DoF pose $\mathbf{T}_k \in \mathcal{SE}(3)$ aligning $\mathcal{P}$ to $\mathcal{Q}_k$, which consists of a 3D rotation $\mathbf{R}_k \in \mathcal{SO}(3)$ and a 3D translation $\mathbf{t}_k \in \mathbb{R}^3$.
To solve this problem, one usually first extracts a set of correspondences $\mathcal{C} \tight{=} \{ (\mathbf{p}_i, \mathbf{q}_i) \tight{\mid} \mathbf{p}_i \tight{\in} \mathcal{P}, \mathbf{q}_i \tight{\in} \mathcal{Q}\}$ between the two point clouds. The correspondences are then clustered into different groups $\{ \mathcal{C}_1, ..., \mathcal{C}_K \}$, and each group recovers the pose of one instance by solving the following problem:
\begin{equation}
\min_{\mathbf{R}_k,\mathbf{t}_k} \sum\nolimits_{(\textbf{p}_i, \textbf{q}_i) \in \mathcal{C}_k} \lVert {\mathbf{R}_k} \cdot \mathbf{p}_{i} + \mathbf{t}_k - \textbf{q}_{i} \rVert^2_2.
\label{eq:procrustes}
\end{equation}
This problem is very challenging as the number of instances is unknown and it is difficult to cluster the correspondences to the correct instances. In this work, we deal with this problem by extracting instance-aware correspondences.


\section{Method}
\label{sec:method}


\begin{figure*}[t]
\begin{overpic}[width=1.0\linewidth]{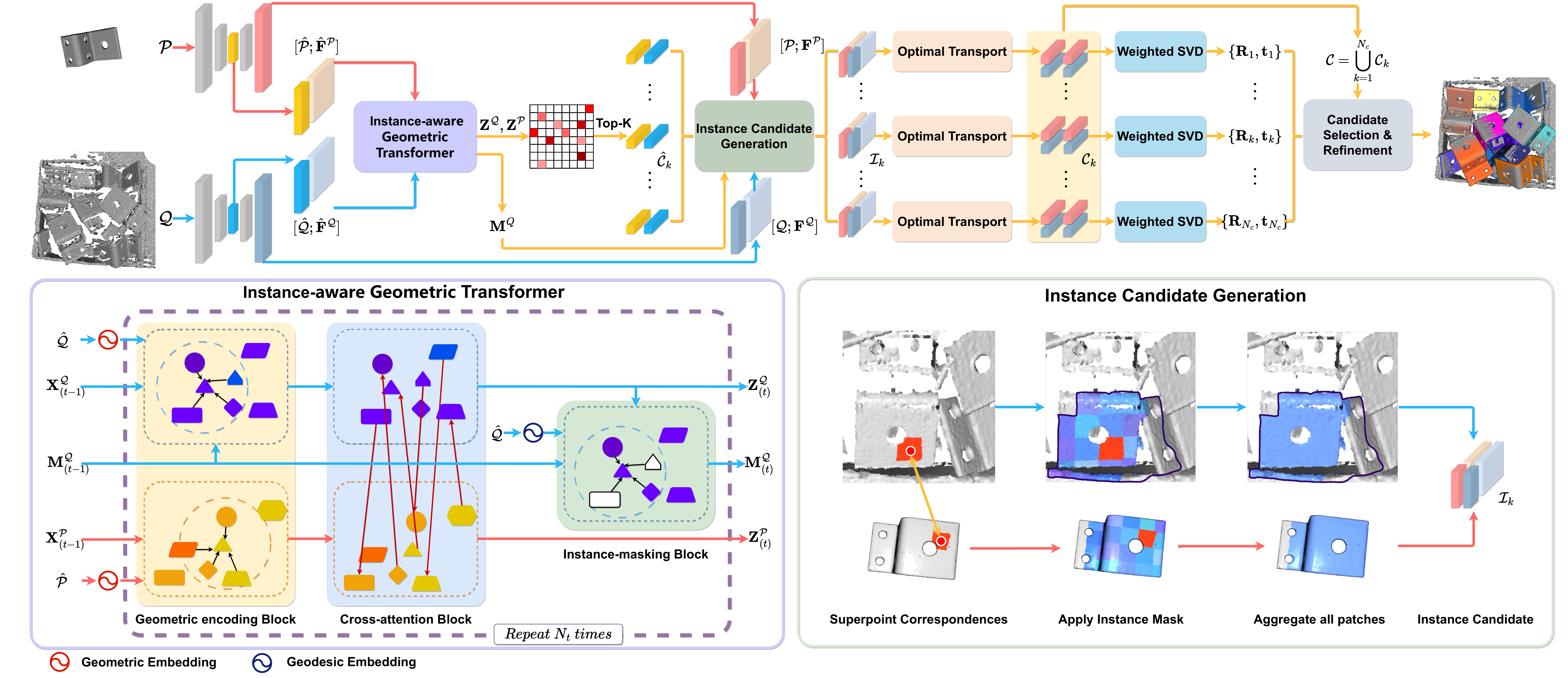}
\end{overpic}
\vspace{-20pt}
\caption{Overall pipeline of \ours{}. The backbone progressively downsamples two point clouds and extracts multi-level features. At the coarse level, the Instance-aware Geometric Transformer module extracts instance-aware superpoint features and establishes reliable superpoint correspondences. At the fine level, the superpoint correspondences are extended to instance candidates, where instance-wise point correspondences are extracted to estimate per-candidate poses. At last, a simple but effective candidate selection and refinement algorithm is adopted to generate the final registrations.}
\label{fig:pipline}
\vspace{-15pt}
\end{figure*}

\ours{} extracts correspondences in a coarse-to-fine fashion similar to keypoint-free registration methods.
The overall pipeline of our method is illustrated in \cref{fig:pipline}.
 At the coarse level, we establish correspondences between downsampled superpoints with an \emph{Instance-aware Geometric Transformer} module~(\cref{sec:model-pam}).
At the fine level, each superpoint correspondence is extended to form an instance candidate, where \emph{instance-wise point correspondences} are extracted for pose estimation (\cref{sec:model-pom}).
At last, we merge similar instance candidates and refine the resultant poses to obtain the final registrations (\cref{sec:model-cluster}).

Given the input point clouds $\mathcal{P}$ and $\mathcal{Q}$, we progressively downsample them into sparse superpoints, denoted as $\hat{\mathcal{P}}$ and $\hat{\mathcal{Q}}$, using grid subsampling~\cite{thomas2019kpconv}. KPFCNN~\cite{thomas2019kpconv} is adopted to extract multi-level point features. Let us denote the features of $\mathcal{P}$, $\mathcal{Q}$, $\hat{\mathcal{P}}$ and $\hat{\mathcal{Q}}$ as $\mathbf{F}^{\mathcal{P}}$, $\mathbf{F}^{\mathcal{Q}}$, $\hat{\mathbf{F}}^{\mathcal{P}}$ and $\hat{\mathbf{F}}^{\mathcal{Q}}$, respectively. Following~\cite{yu2021cofinet,qin2022geometric}, each superpoint $\hat{\mathbf{p}}_i \tight{\in} \hat{\mathcal{P}}$ ($\hat{\mathbf{q}}_i \tight{\in} \hat{\mathcal{Q}}$) is associated with a local patch $\mathcal{G}^{\mathcal{P}}_i$ ($\mathcal{G}^{\mathcal{Q}}_i$) by the point-to-node partition strategy~\cite{li2018so} where each point is assigned to its nearest superpoint.

\subsection{Instance-aware Geometric Transformer}
\label{sec:model-pam}
\begin{figure}[t]
\begin{overpic}[width=1.0\linewidth]{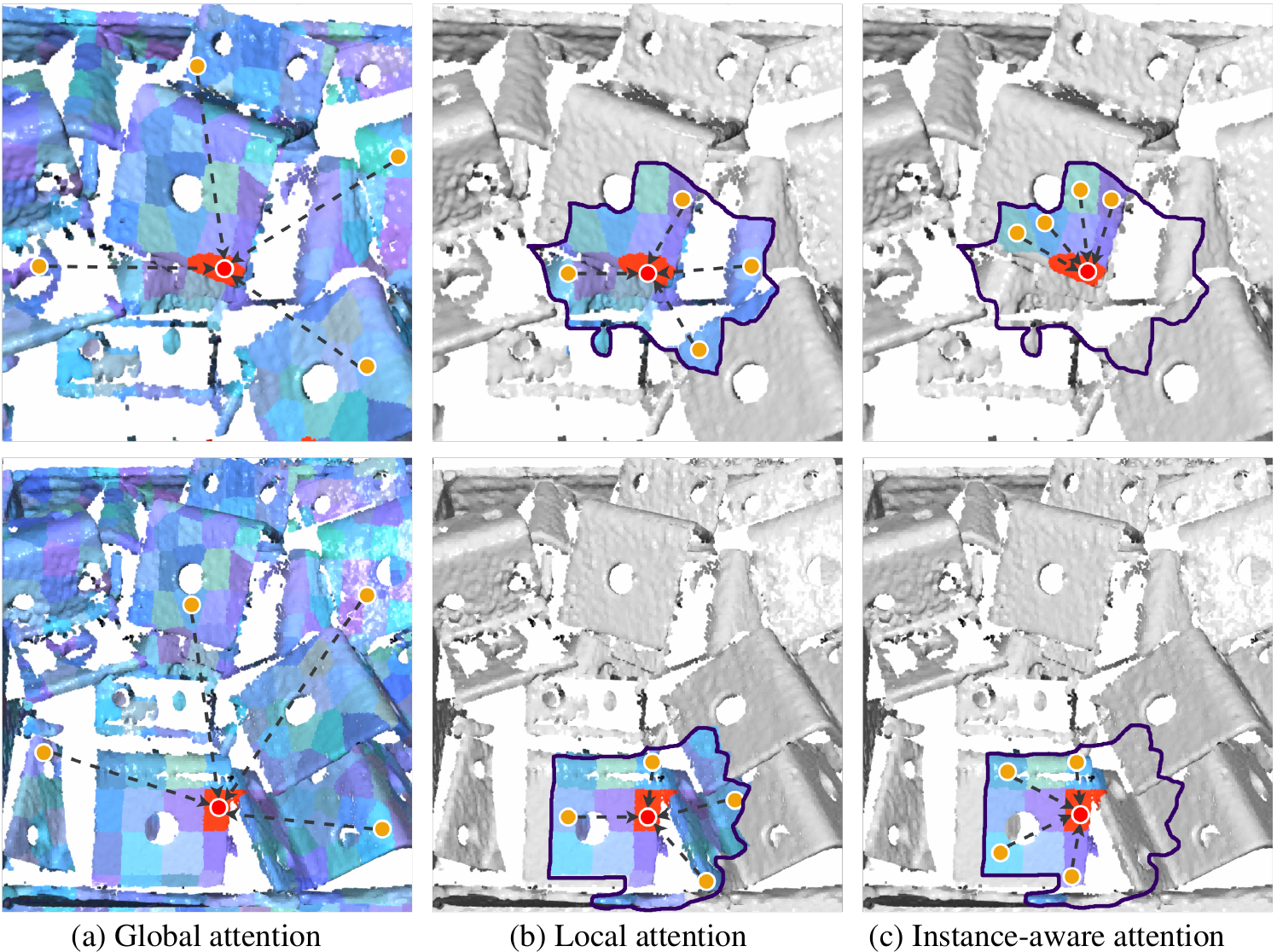}
\end{overpic}
\vspace{-20pt}
\caption{Comparison of (a) global attention, (b) local attention, and (c) instance-aware attention. The superpoints (patches) participating in the attention computation are color-coded. The anchor superpoints are in \textcolor{red}{red}. The $k$-nearest neighbors of the anchor are enclosed by the \textcolor{DeepPurple}{purple} line.}
\label{fig:different-attention}
\vspace{-15pt}
\end{figure}

Given the superpoints $\hat{\mathcal{P}}$ and $\hat{\mathcal{Q}}$, as well as their features $\hat{\mathbf{F}}^{\mathcal{P}}$ and $\hat{\mathbf{F}}^{\mathcal{Q}}$, we first extract a set of superpoint correspondences whose local patches overlap with each other.
Accurate superpoint matching relies on learning the geometric consistency between the two point clouds.
To this end, previous works~\cite{yu2021cofinet,qin2022geometric} adopt transformer~\cite{vaswani2017attention} to encode intra- and inter-point-cloud context in the global scope, as shown in \cref{fig:different-attention}~(a).
Albeit successful in pairwise registration, this is problematic in multi-instance registration.
In the scene point cloud, the superpoints from the background or other instances are de facto noises for modeling the geometric consistency and severely pollute the superpoint features.
This becomes more severe in cluttered scenes.
Considering an heavily-occluded instance, as its geometric structure is incomplete, the features of its superpoints could be easily overwhelmed by those from the background or nearby instances. As a result, it can hardly be registered.

To address these issues, we propose to make the superpoint features \emph{instance-aware} with a novel \emph{Instance-aware Geometric Transformer} module. The key insight is to restrict the intra-point-cloud context encoding in the scene point cloud within each individual instance.
To this end, we first extract the $k$-nearest neighbors for each superpoint and conduct context aggregation within local regions.
However, as shown in \cref{fig:different-attention}~(b), nearby superpoints do not necessarily reside in the same instance, and the feature pollution problem still exists.
For this reason, we further design an instance masking mechanism by predicting an instance mask for each superpoint to select the neighbors in the same instance as it, and only aggregate the context among them.
\cref{fig:different-attention}~(c) illustrates our instance masking mechanism.

As shown in \cref{fig:pipline}~(bottom), the instance-aware geometric transformer contains three blocks.
First, a \emph{geometric encoding block} encodes intra-instance geometric structure based on the instance masks.
Next, a \emph{cross-attention block} enhances the superpoint features by modeling inter-point-cloud geometric consistency.
At last, an \emph{instance masking block} predicts new instance masks with the instance-aware superpoint features.
We adopt $N_t$ instance-aware geometric transformer modules to progressively refine the superpoint features and the instance masks.

\ptitle{Geometric encoding block.}
This block encodes the intra-instance geometric context within each point cloud.
Given an anchor superpoint $\hat{\mathbf{q}_i} \in \hat{\mathcal{Q}}$, its $k$-nearest superpoints $\mathcal{N}^{\mathcal{Q}}_i = \{\hat{\mathbf{q}}_{i_1}, ..., \hat{\mathbf{q}}_{i_k}\}$, the input feature matrix $\mathbf{X}^{\mathcal{Q}} \in \mathbb{R}^{\lvert \hat{\mathcal{Q}} \rvert \times d}$, and the instance mask matrix $\mathbf{M}^{\mathcal{Q}} \in \mathbb{R}^{\lvert \hat{\mathcal{Q}} \rvert \times k}$, the output feature matrix $\mathbf{Z}^{\mathcal{Q}} \in \mathbb{R}^{\lvert \hat{\mathcal{Q}} \rvert \times d}$ is computed as:
\begin{equation}
\mathbf{z}^{\mathcal{Q}}_i = \sum_{j=1}^{k} \frac{\exp(e_{i, j})}{\sum_{l=1}^{k} \exp(e_{i, l})} (\mathbf{x}^{\mathcal{Q}}_{i_j} \mathbf{W}^V),
\label{eq:geometric-features}
\end{equation}
where the attention score $e_{i, j}$ is computed as:
\begin{equation}
e_{i, j} = \frac{(\mathbf{x}^{\mathcal{Q}}_i \mathbf{W}^Q)(\mathbf{x}^{\mathcal{Q}}_{i_j} \mathbf{W}^K + \mathbf{r}_{i, j} \mathbf{W}^R)^{\top}}{\sqrt{d}} + m^{\mathcal{Q}}_{i, j},
\label{eq:geometric-attention}
\end{equation}
where $\mathbf{W}^Q, \mathbf{W}^K, \mathbf{W}^V, \mathbf{W}^R \in \mathbb{R}^{d \times d}$ are the projection weights for query, key, value and geometric embedding, respectively.
$\mathbf{r}_{i, j}$ is the geometric structure embedding~\cite{qin2022geometric} encoding the transformation-invariant geometric information among the superpoints.
For the instance mask $\mathbf{M}^{\mathcal{Q}}$, we set $m^{\mathcal{Q}}_{i, j} = 0$ if $\hat{\mathbf{q}}_i$ and $\hat{\mathbf{q}}_{i_j}$ are in the same instance, while $m^{\mathcal{Q}}_{i, j} = -\infty$ otherwise. $\mathbf{M}^{\mathcal{Q}}$ is initialized with all zeros and refined by the instance masking block described later. Benefiting from the instance masks, the superpoint features are instance-aware by encoding the feature correlation and the geometric context within their belonging instances.
For the computation in $\hat{\mathcal{P}}$, we ignore the mask term in \cref{eq:geometric-attention} and aggragate the features from all neighbors.


\ptitle{Cross-attention block.}
After encoding the intra-instance geometric context, we further learn the inter-point-cloud geometric consistency with a cross-attention block inspired by~\cite{yu2021cofinet,qin2022geometric}. Given the feature matrices $\mathbf{X}^{\mathcal{P}} \tight{\in} \mathbb{R}^{\lvert \hat{\mathcal{P}} \rvert \times d}$ of $\hat{\mathcal{P}}$ and $\mathbf{X}^{\mathcal{Q}} \tight{\in} \mathbb{R}^{\lvert \hat{\mathcal{Q}} \rvert \times d}$ of $\hat{\mathcal{Q}}$ from the geometric encoding block, the output feature matrix $\mathbf{Z}^{\mathcal{P}} \tight{\in} \mathbb{R}^{\lvert \hat{\mathcal{P}} \rvert \times d}$ of $\hat{\mathcal{P}}$ is computed as:
\begin{equation}
\mathbf{z}^{\mathcal{P}}_i = \sum_{j=1}^{\lvert \hat{\mathcal{Q}} \rvert} \frac{\exp(e_{i, j})}{\sum_{k=1}^{\lvert \hat{\mathcal{Q}} \rvert} \exp(e_{i, k})} (\mathbf{x}^{\mathcal{Q}}_{k} \mathbf{W}^V),
\end{equation}
where the attention score $e_{i, j}$ is computed as:
\begin{equation}
e_{i, j} = \frac{(\mathbf{x}^{\mathcal{P}}_i \mathbf{W}^Q)(\mathbf{x}^{\mathcal{Q}}_{j} \mathbf{W}^K)^{\top}}{\sqrt{d}},
\end{equation}
where $\mathbf{W}^Q, \mathbf{W}^K, \mathbf{W}^V \in \mathbb{R}^{d \times d}$ are the respective projection weights for query, key and value.
The same computation goes for $\mathbf{Z}^{\mathcal{Q}}$.
Benefiting from the cross-attention block, the superpoint features in one point cloud are aware of the geometric structure of the other one, which facilitates modeling the geometric consistency between two point clouds.


\begin{figure}[t]
\begin{overpic}[width=1.0\linewidth]{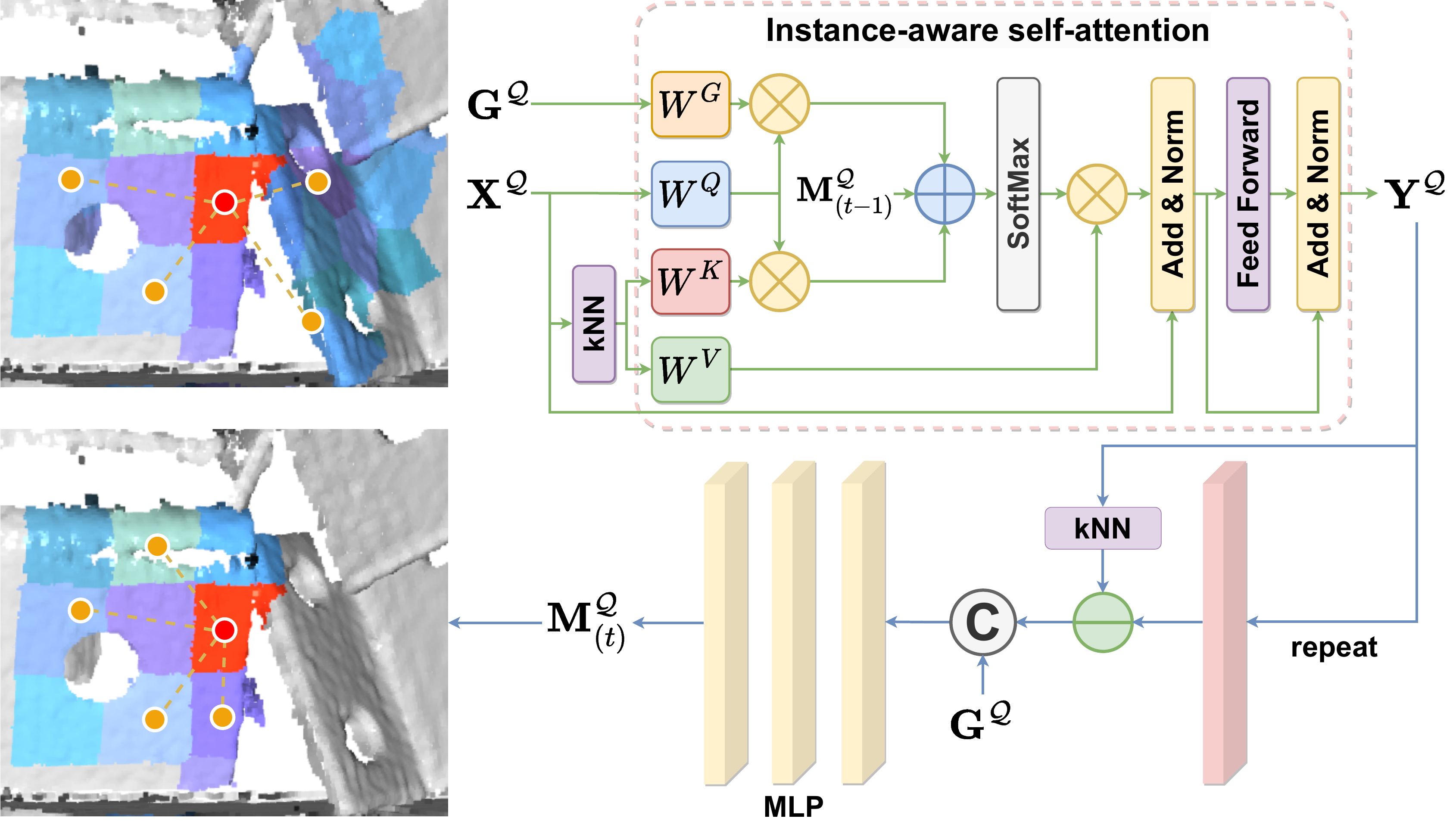}
\end{overpic}
\vspace{-20pt}
\caption{Structure of the Instance masking block.}
\label{fig:attention}
\vspace{-15pt}
\end{figure}

\ptitle{Instance masking block.}
At last, this block consumes the cross-attention features $\mathbf{X}^{\mathcal{Q}}$ of $\hat{\mathcal{Q}}$ as well as the previous instance mask $\mathbf{M}^{\mathcal{Q}}_{(t-1)}$ to predict a refined instance mask $\mathbf{M}^{\mathcal{Q}}_{(t)}$.
We observe that two superpoints from different instances tend to be distant in the geodesic space.
To this end, we design a geodesic self-attention mechanism by replacing the geometric structure embedding in \cref{eq:geometric-attention} with a similar geodesic distance embedding to enhance the superpoint features.
Please refer to the appendix for more details.
The resultant superpoint features are denoted as $\mathbf{Y}$.
At last, we adopt an MLP to predict a confidence score $u_{i, j}$ for each neighbor $\hat{\mathbf{q}}_{i_j} \tight{\in} \mathcal{N}^{\mathcal{Q}}_i$ indicating whether it belongs to the same instance as $\hat{\mathbf{q}}_{i}$ based on the feature discrepancy between $\mathbf{y}_i$ and $\mathbf{y}_{i_j}$ and the geodesic distance embedding $\mathbf{g}_{i, j}$ between $\hat{\mathbf{q}}_i$ and $\hat{\mathbf{q}}_{i_j}$:
\begin{equation}
u_{i, j} = \sigma\Bigl(\mathtt{MLP}\bigl([\hspace{3pt}\mathbf{y}_{i_j} - \mathbf{y}_i;\hspace{3pt}\mathbf{g}_{i, j}\hspace{3pt}]\bigr)\Bigr),
\end{equation}
where $\sigma(\cdot)$ is the sigmoid function and $[\cdot]$ is the concatenation operator. At last, the confidence matrix $\mathbf{U} \in \mathbb{R}^{\lvert \hat{\mathcal{Q}} \rvert \times k}$ is converted into $\mathbf{M}^{\mathcal{Q}}_{(t)}$ with a thresholding function:
\begin{equation}
m^{\mathcal{Q}}_{i, j} = \begin{cases}
-\infty & u_{i, j} < \tau \\
0 & \text{otherwise}
\end{cases},
\label{eq:mask}
\end{equation}
where $\tau$ is the confidence threshold. Thanks to the instance mask, our model can effectively learn instance-aware superpoint features and extract accurate superpoint correspondences covering more instances. Furthermore, the instance masks also help extract instance-aware dense point correspondences as described in \cref{sec:model-pom}.

\ptitle{Superpoint matching.}
We extract \emph{superpoint correspondences} $\hat{\mathcal{C}}$ by matching the superpoint features from the last cross-attention block. As in~\cite{qin2022geometric}, we select the top $N_{c}$ superpoint pairs with the highest cosine feature similarity as the superpoint correspondences.

\subsection{Instance Candidate Generation}
\label{sec:model-pom}

After obtaining the superpoint correspondences, we then refine them to dense point correspondences at the fine level.
Previous methods~\cite{yu2021cofinet,qin2022geometric} opt to match the points within the local patches of two matched superpoints with an optimal transport layer~\cite{sarlin2020superglue}. Nonetheless, the local correspondences extracted in this manner often cluster closely, which could lead to unstable pose estimation as noted in~\cite{huang2021predator}. This is aggravated in multi-instance registration, especially in cluttered scenes, as each instance is typically small and thus its pose cannot be globally optimized as in~\cite{qin2022geometric}.

To address this issue, we propose to extract dense point correspondences \emph{within the instance scope} by leveraging the instance masks from the coarse level. For each superpoint correspondence $\hat{\mathcal{C}}_k = (\hat{\mathbf{p}}_i, \hat{\mathbf{q}}_j)$, we collect their neighboring superpoints $\mathcal{N}^{\mathcal{P}}_i$ and $\mathcal{N}^{\mathcal{Q}}_j$. The superpoints from different instances of $\hat{\mathbf{q}}_j$ are removed from $\mathcal{N}^{\mathcal{Q}}_j$ based on the instance mask $\mathbf{M}_j$.
The points in the local patches of all superpoints in $\mathcal{N}^{\mathcal{P}}_i$ and $\mathcal{N}^{\mathcal{Q}}_j$ imply a potential occurrence of $\mathcal{P}$ in $\mathcal{Q}$, \ie, an \emph{instance candidate}, denoted as $\mathcal{I}_k$. 
Next, instance-wise \emph{point correspondences} are extracted within $\mathcal{I}_k$ with an optimal transport layer and mutual top-$k$ selection following~\cite{qin2022geometric}, denoted as $\mathcal{C}_k$.
At last, we estimate a pose $\mathbf{T}_k = \{\mathbf{R}_k, \mathbf{t}_k\}$ for $\mathcal{I}_k$ by solving \cref{eq:procrustes} with weighted SVD~\cite{besl1992method}.
Thanks to the powerful instance-aware geometric transformer, an instance candidate can cover a relatively large portion of an instance, with few points from the background or other instances, so the poses obtained in this step have already been very accurate.

\subsection{Candidate Selection and Refinement}
\label{sec:model-cluster}

As several superpoint correspondences could belong to the same instance, there are commonly duplicated instances in the instance candidates. For this reason, we further design a simple but effective algorithm for duplicate removal and candidate refinement.

Inspired by non-maximum suppresion (NMS)~\cite{rothe2015non}, we first sort the instance candidates by the inlier ratio on the global point correspondences $\mathcal{C} = \bigcup_{i=1}^{N_c} \mathcal{C}_i$. The inlier ratio of the instance candidate $\mathcal{I}_k$ is:
\begin{equation}
\gamma_k = \frac{1}{\lvert \mathcal{C} \rvert} \sum_{(\mathbf{p}, \mathbf{q}) \in \mathcal{C}} \llbracket \lVert \mathbf{R}_k \mathbf{p} + \mathbf{t}_k - \mathbf{q} \rVert < \tau_2 \rrbracket,
\end{equation}
where $\llbracket \cdot \rrbracket$ is the Iverson bracket and $\tau_2$ is the acceptance radius.
Next, we select the instance candidate with the highest inlier ratio as the anchor candidate.
All remaining candidates similar with the anchor are merged with it and removed from future computation.
We define the similarity between two candidates $\mathcal{I}_i$ and $\mathcal{I}_j$ by the average distance (ADD)~\cite{hinterstoisser2013model} between their estimated poses:
\begin{equation}
s_{i, j} = 1 - \frac{\mathtt{ADD}(\mathbf{T}_i, \mathbf{T}_j)}{r},
\end{equation}
where $r$ is the diameter of $\mathcal{P}$.
And $\mathcal{I}_i$ and $\mathcal{I}_j$ are considered similar if $s_{i, j}$ is above $\tau_s$.
To merge two candidates, we combine their point correspondences and recompute the pose by solving \cref{eq:procrustes}.
At last, the preserved pose is iteratively refined with the surviving inliers as in~\cite{bai2021pointdsc,qin2022geometric}, leading to a final instance registration.
The anchor is then removed from future computation.
The above process repeats until there are no candidates left and we omit the instance registrations with too few inliers.

Compared to multi-model fitting methods~\cite{tang2022multi,yuan2022pointclm}, the advantages of our \ours{} are two-fold. First, by learning instance-aware superpoint features, the influence from the background and other instances is minimized, thus the correspondences are more accurate and can cover more instances. Second, correspondence clustering-based methods tend to lose heavily-occluded instances as there are few correspondences on them, especially in cluttered scenes. On the contrary, our method can better recognize these hard instances as we bypass the clustering process by directly extracting instance candidates and merging similar ones.

\subsection{Loss Functions}
\label{sec:model-loss}

We adopt three loss functions to train \ours{}: an overlap-aware circle loss~\cite{qin2022geometric} $\mathcal{L}_{\text{circle}}$ to supervise the superpoint features, a negative log-likelihood loss~\cite{sarlin2020superglue} $\mathcal{L}_{\text{nll}}$ to supervise the point matching, and a mask prediction loss~\cite{milletari2016v} $\mathcal{L}_{\text{mask}}$ to supervise the instance masks.
The overall loss is computed as $\mathcal{L} = \mathcal{L}_{\text{circle}} + \mathcal{L}_{\text{nll}} + \mathcal{L}_{\text{mask}}$. Please refer to the supplementary material for more details.


\section{Experiments}
\label{sec:experiments}

We evaluate \ours{} on indoor Scan2CAD~\cite{avetisyan2019scan2cad,tang2022multi,yuan2022pointclm}, industrial ROBI~\cite{yang2021robi} and synthetic ShapeNet~\cite{chang2015shapenet} benchmarks. And we also introduce the implementation details and more experiments in the supplementary material.

\ptitle{Baselines.}
We compare with three state-of-the-art point cloud correspondence methods, FCGF~\cite{choy2019fully}, CoFiNet~\cite{yu2021cofinet} and GeoTransformer~\cite{qin2022geometric}, and four recent multi-model fitting methods, T-linkage~\cite{magri2014t}, RansaCov~\cite{magri2016multiple}, ECC~\cite{tang2022multi}, and PointCLM~\cite{yuan2022pointclm}. The point cloud correspondence methods and the multi-model fitting methods are pairwise integrated for a comprehensive comparison.

\ptitle{Metrics.} 
Following~\cite{tang2022multi,yuan2022pointclm}, we evaluate our method with three registration metrics: (1) \emph{Mean Recall} (MR), the ratio of registered instances over all ground-truth instances, (2) \emph{Mean Precision} (MP), the ratio of registered instances over all predicted instances, and (3) \emph{Mean F$_1$ score} (MF), the harmonic mean of MP and MR. We also report \emph{Inlier Ratio} (IR), the ratio of inliers over all extracted correspondences.

\subsection{Evaluations on Scan2CAD}
\label{sec:exp-scan2cad}

\ptitle{Dataset.}
Scan2CAD~\cite{avetisyan2019scan2cad} is a scene-to-CAD alignment dataset build upon ScanNet~\cite{dai2017scannet} and ShapeNet~\cite{chang2015shapenet}. It consists of $1506$ scenes from ScanNet annotated with $14225$ CAD models from ShapeNet and their poses in the scenes. Following~\cite{yuan2022pointclm}, we replace the objects in the scenes with the corresponding aligned CAD models. We select the scene-model pairs which the scene contain multiple instances of the model for multi-instance registration. As last, we obtain $2184$ point cloud pairs and use $70\%$ pairs for training, $10\%$ for validation and $20\%$ for testing.


\begin{table}[!t]
\centering
\scriptsize
\setlength{\tabcolsep}{4pt}
\begin{tabular}{l|c|ccc}
\toprule
Model & IR($\%$) & MR($\%$) & MP($\%$) & MF($\%$) \\
\midrule
FCGF~\cite{choy2019fully} + T-Linkage~\cite{magri2014t} & \multirow{4}{*}{36.27} & 54.75 & 22.76 & 32.15 \\
FCGF~\cite{choy2019fully} + RansaCov~\cite{magri2016multiple} &  & 73.50 & 45.01 & 55.83 \\
FCGF~\cite{choy2019fully} + PointCLM~\cite{yuan2022pointclm} &  & 86.99 & 70.05 & 77.60 \\
FCGF~\cite{choy2019fully} + ECC~\cite{tang2022multi} &  & 92.60 & 73.79 & 82.13 \\
\midrule
CoFiNet~\cite{yu2021cofinet} + T-Linkage~\cite{magri2014t} & \multirow{4}{*}{18.52} & 28.17 & 5.09 & 8.62 \\
CoFiNet~\cite{yu2021cofinet} + RansaCov~\cite{magri2016multiple} &  & 38.29 & 13.69 & 20.16 \\
CoFiNet~\cite{yu2021cofinet} + PointCLM~\cite{yuan2022pointclm} &  & 36.15 & 24.70 & 29.34 \\
CoFiNet~\cite{yu2021cofinet} + ECC~\cite{tang2022multi} &  & 57.26 & 17.24 & 26.50 \\   
\midrule
GeoTransformer~\cite{qin2022geometric} + T-Linkage~\cite{magri2014t} & \multirow{4}{*}{\underline{44.02}} & 72.68 & 44.80 & 55.43 \\
GeoTransformer~\cite{qin2022geometric} + RansaCov~\cite{magri2016multiple} &  & 78.84 & 67.16 & 72.53 \\
GeoTransformer~\cite{qin2022geometric} + PointCLM~\cite{yuan2022pointclm} &  & 82.81 & 81.90 & 82.35 \\
GeoTransformer~\cite{qin2022geometric} + ECC~\cite{tang2022multi} &  & 94.63 & 74.83 & 83.57 \\
\midrule
\ours{} (\emph{ours}) + T-Linkage~\cite{magri2014t} & \multirow{5}{*}{\textbf{56.59}} & 77.12 & 46.04 & 57.65 \\
\ours{} (\emph{ours}) + RansaCov~\cite{magri2016multiple} &  & 84.78 & 71.34 & 77.48 \\
\ours{} (\emph{ours}) + PointCLM~\cite{yuan2022pointclm} &  & 91.85 & \underline{91.08} & \underline{91.46} \\
\ours{} (\emph{ours}) + ECC~\cite{tang2022multi} &  & \textbf{96.52} & 89.03 & 92.62 \\
\ours{} (\emph{ours}, full pipeline) &  & \underline{95.70} & \textbf{91.21} & \textbf{93.40} \\ 
\bottomrule
\end{tabular}
\vspace{-5pt}
\caption{Evaluation results on Scan2CAD benchmark.}
\label{table:results-scancad}
\vspace{-15pt}
\end{table}

\ptitle{Quantitative results.}
\label{sec:exp-scan2cad-registration}
As shown in \cref{table:results-scancad}, our \ours{} achieves consistent improvements over the baseline correspondence models. Notably, our method surpasses the previous best GeoTransformer on IR by $12$ percentage points (pp). When coupled with PointCLM, it outperforms GeoTransformer by over $9$ pp on the three registration metrics.

Moreover, our full pipeline surpasses all models with the baseline multi-model fitting methods. Although the ECC-based model performs slightly better than full \ours{} on MR, its MP is significantly worse. This indicates that ECC achieves high MR by predicting more registrations, which is impractical in real-world applications. And our method outperforms PointCLM on all three registration metrics, which demonstrates the strong effectiveness of our method.


\begin{figure*}[t]
\begin{overpic}[width=1.0\linewidth]{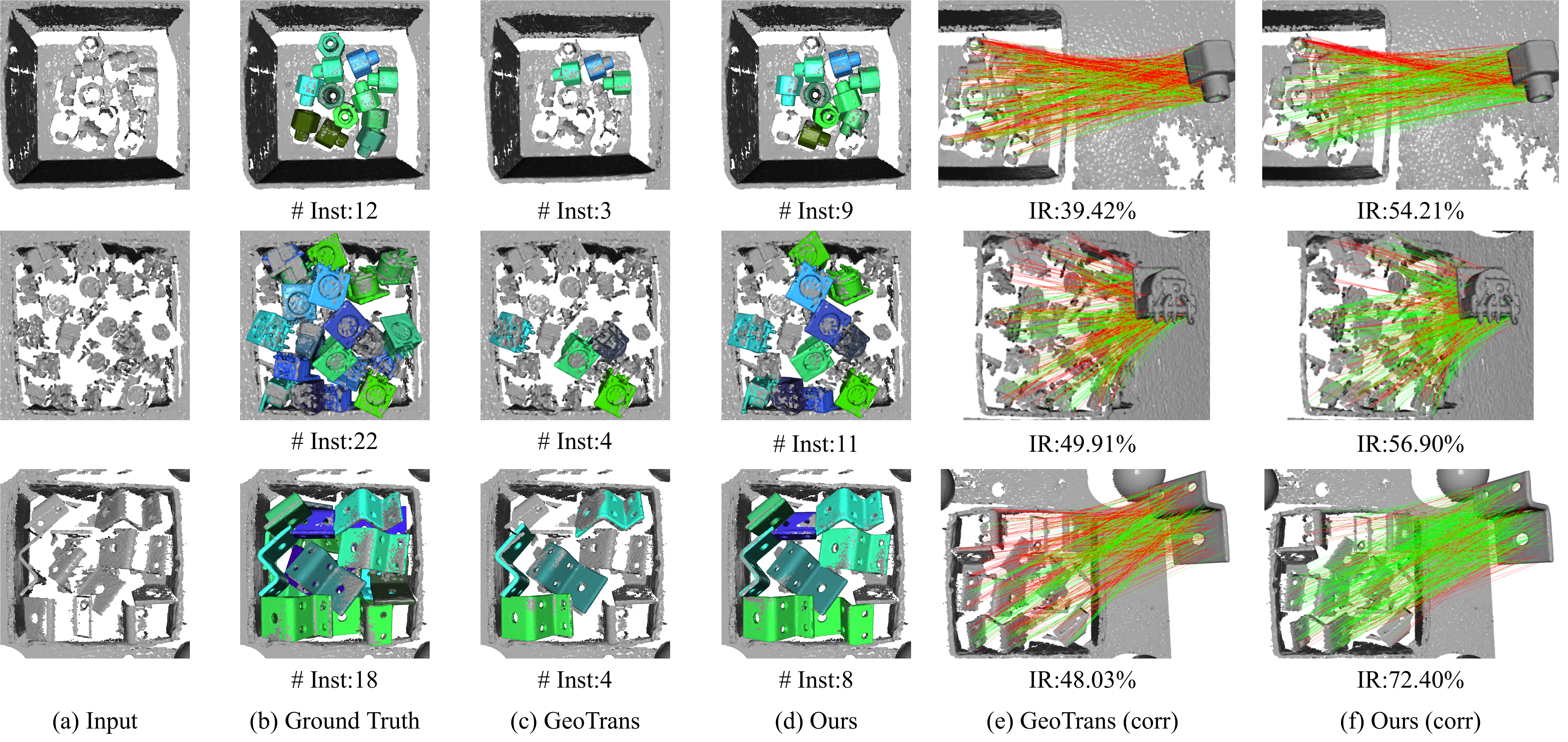}
\end{overpic}
\vspace{-20pt}
\caption{Registration results on ROBI benchmark. We visualize the successfully registered instances in (c) and (d). \ours{} registers more instances in the cluttered scenes (the $2^{\text{nd}}$ and the $3^{\text{rd}}$ rows) and with incomplete geometry (all three rows).
And it extracts more accurate correspondences benefiting from the instance-aware correspondence learning mechanism.
}
\label{fig:robi}
\vspace{-15pt}
\end{figure*}


\subsection{Evaluations on ROBI}
\label{sec:exp-robi}

\ptitle{Dataset.}
ROBI~\cite{yang2021robi} is a recent dataset for industrial bin-picking. It includes $7$ reflective metallic industrial objects and $63$ bin-picking scenes. Each scene contains cluttered instances of one industrial object.
For each point cloud pair, the scene point cloud is backprojected from a depth image and the model point cloud is sampled from the CAD model of its corresponding industrial object.
We obtain $4880$ pairs in total, and we split the scenes for training, validation and testing according to the ratio of $6:1:2$ for each object.


\begin{table}[!t]
\centering
\scriptsize
\setlength{\tabcolsep}{4pt}
\begin{tabular}{l|c|ccc}
\toprule
Model & IR($\%$) & MR($\%$) & MP($\%$) & MF($\%$) \\
\midrule
CoFiNet~\cite{yu2021cofinet} + T-Linkage~\cite{magri2014t} & \multirow{4}{*}{10.35} & 1.34 & 0.72 & 0.93 \\
CoFiNet~\cite{yu2021cofinet} + RansaCov~\cite{magri2016multiple} &  & 1.25 & 1.39 & 1.31 \\
CoFiNet~\cite{yu2021cofinet} + PointCLM~\cite{yuan2022pointclm} &  & 1.17 & 2.11 & 1.50 \\
CoFiNet~\cite{yu2021cofinet} + ECC~\cite{tang2022multi} &  & 3.21 & 8.50 & 4.66 \\
\midrule
GeoTransformer~\cite{qin2022geometric} + T-Linkage~\cite{magri2014t} & \multirow{4}{*}{\underline{39.64}} & 7.71  &4.57 & 5.73 \\
GeoTransformer~\cite{qin2022geometric} + RansaCov~\cite{magri2016multiple} &  & 8.99 & 13.58 & 10.81 \\
GeoTransformer~\cite{qin2022geometric} + PointCLM~\cite{yuan2022pointclm} &  & 13.98 & 26.64 & 18.33 \\
GeoTransformer~\cite{qin2022geometric} + ECC~\cite{tang2022multi} &  & 18.55 & 30.99 & 
23.20 \\
\midrule
\ours{} (\emph{ours}) + T-Linkage~\cite{magri2014t} & \multirow{5}{*}{\textbf{45.14}} & 12.04& 10.47 & 11.20 \\ 
\ours{} (\emph{ours}) + RansaCov~\cite{magri2016multiple} &  &14.14 & 26.29 & 18.38 \\ 
\ours{} (\emph{ours}) + PointCLM~\cite{yuan2022pointclm} &  & 18.68 & \underline{40.11} & \underline{25.48} \\
\ours{} (\emph{ours}) + ECC~\cite{tang2022multi} &  & \underline{24.65} & 34.85 & 28.91 \\
\ours{} (\emph{ours}, full pipeline) &  & \textbf{38.51} & \textbf{41.19} & \textbf{39.80} \\ 
\bottomrule
\end{tabular}
\vspace{-5pt}
\caption{Evaluation results on ROBI benchmark.}
\label{table:results-robi}
\vspace{-18pt}
\end{table}

\ptitle{Quantitative results.}
As in \cref{table:results-robi}, our method outperforms the baseline correspondence methods by a large margin on all four metrics. Compared to Scan2CAD, ROBI is more challenging due to the more cluttered scenes, where instance-aware information is more important. Benefiting from the instance-aware geometric transformer module, our method extracts reliable instance-aware correspondences, which contributes to more accurate instance registrations.

Our full model achieves significant improvements over the ECC-based model on MR, MP and MF. The PointCLM-based model achieves comparable MP with the full model, but significantly worse MR, which means it can register only several easy instances but misses most heavily-occluded instances. Specifically, it can register on average only $6.2$ instances per scene, while the number of our method is $13.7$. Our method achieves both high MR and MP, showing strong robustness in cluttered scenes.


\begin{figure}[t]
\begin{overpic}[width=1.0\linewidth]{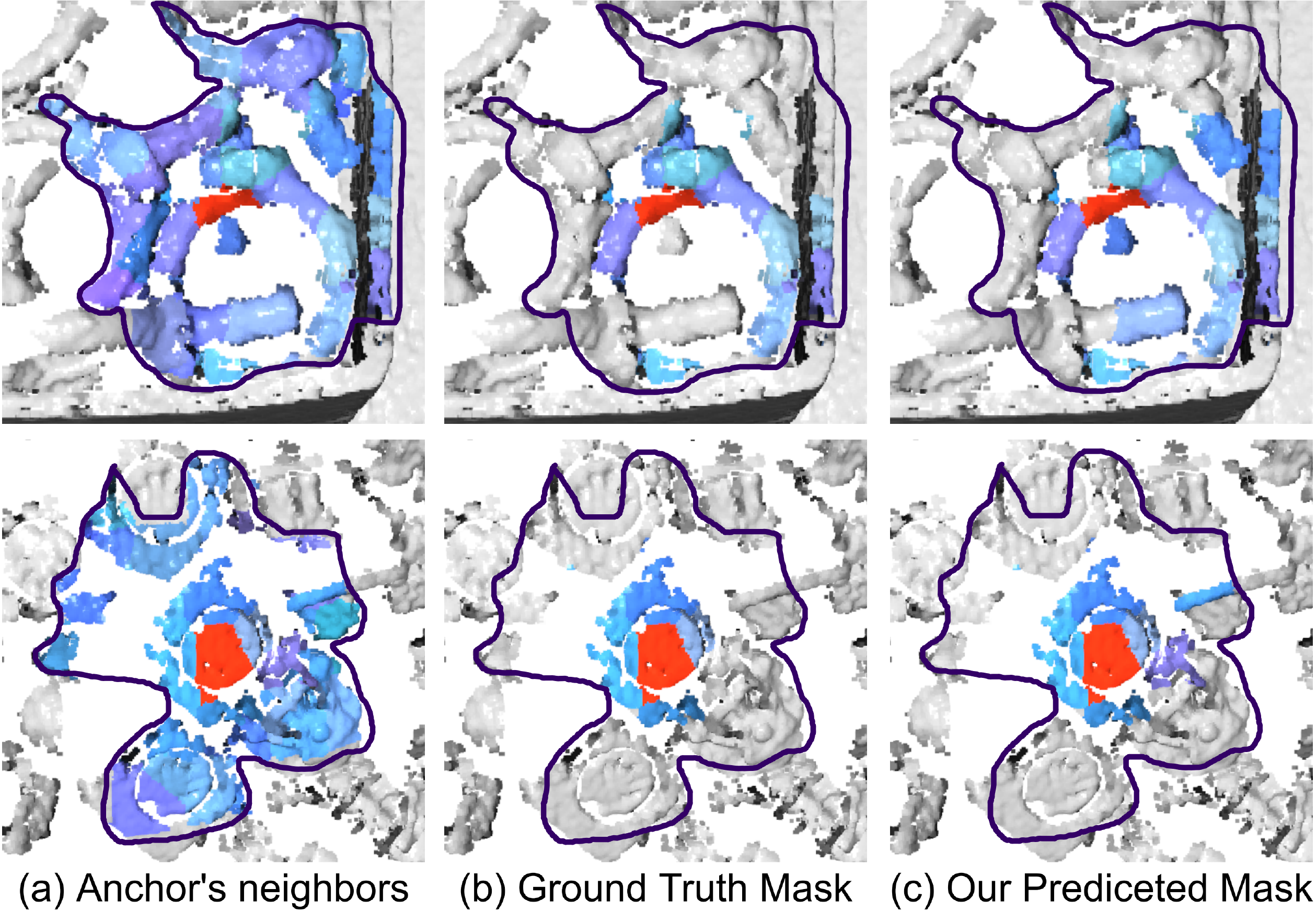}
\end{overpic}
\vspace{-20pt}
\caption{Visualizations of the predicted instance masks. The anchor superpoint (patch) is in \textcolor{red}{red} and the selected neighbors are color-coded. Our method can effectively reject the patches outside the interested instance while preserving most of those inside.}
\label{fig:mask}
\vspace{-18pt}
\end{figure}

\ptitle{Qualitative results.}
\cref{fig:robi} visualizes the registration results of GeoTransformer and \ours{}. Our method can register more instances in cluttered scenes (see the $2^{\text{nd}}$ and the $3^{\text{rd}}$ rows) and with severe geometric deficiency (see all three rows).
And \ours{} attains more high-quality correspondences thanks to our instance-aware design.

We further visualize the predicted instance masks \cref{fig:mask}. The instance masks effectively reject the patches out of the interested instance while preserving most of those inside. Specifically, our method achieves the instance mIoU of $69.26$ pp while the result of the model without masking is merely $38.13$ pp.
The high-quality masks allow us to effectively encode intra-instance context and contribute to accurate registrations.

\begin{figure}[t]
\begin{overpic}[width=1.0\linewidth]{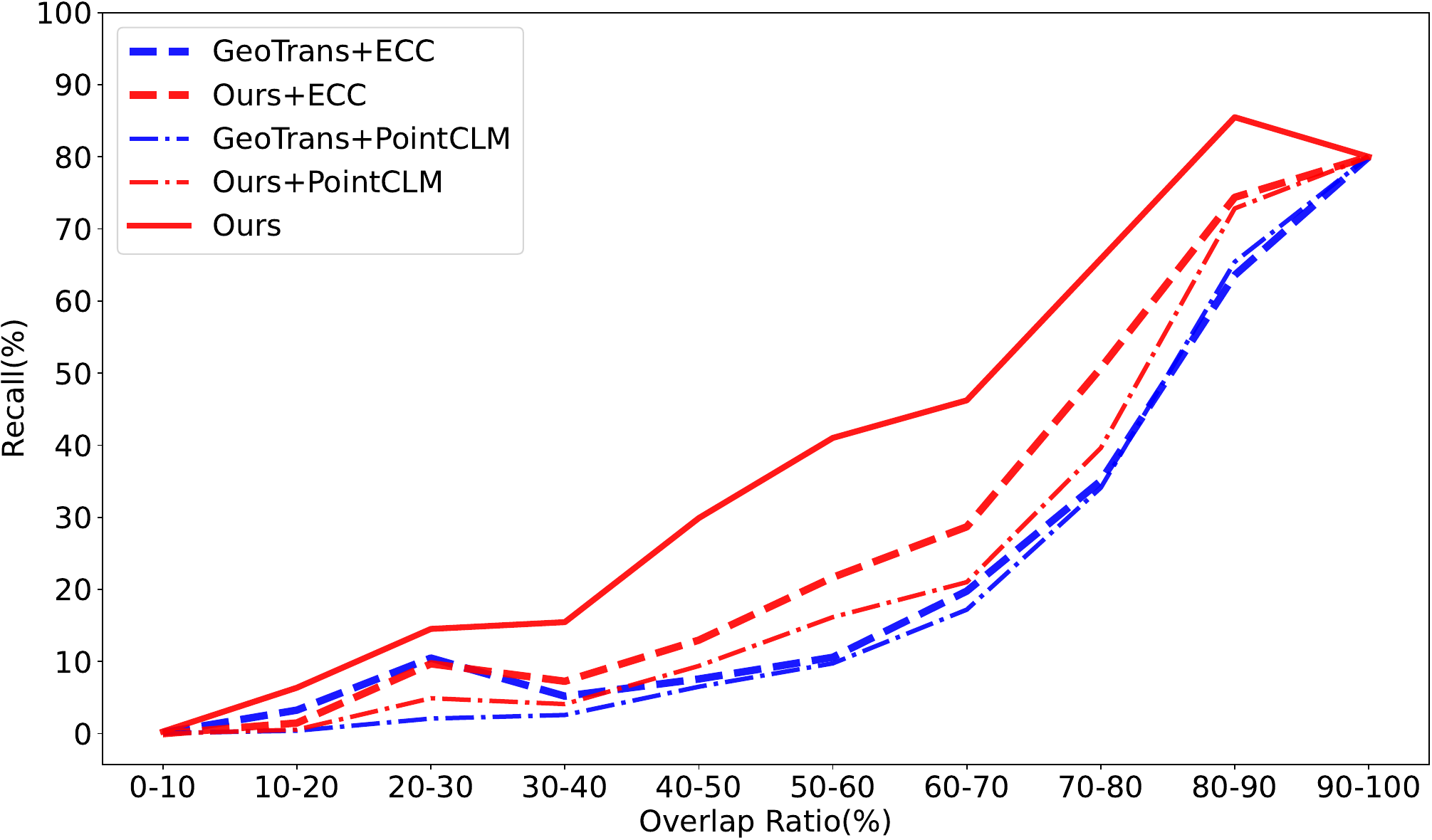}
\end{overpic}
\vspace{-20pt}
\caption{Results in different overlap levels on ROBI benchmark.}
\label{fig:robi_overlap}
\vspace{-15pt}
\end{figure}

\cref{fig:robi_overlap} compares the results under different overlap ratios. \ours{} consistently outperforms the baselines at all overlap levels, and the improvements are more significant when the overlap ratio is below $50\%$. This proves the strong capability of our method to handle low-overlap instances.


\subsection{Evaluations on ShapeNet}
\label{sec:exp-ShapeNet}

\ptitle{Dataset.}
ShapeNet~\cite{chang2015shapenet} is a large CAD model dataset which contains $51300$ models from $55$ categories. We use the models from first $30$ categories for training and those from the rest $25$ categories for testing to evaluate the generality to novel categories. We randomly sample at most $500$ models from each category to avoid class imbalance. For each point cloud pair, the model point cloud is a random CAD model and the scene one is constructed by applying $4 {\sim} 16$ random poses on the model. At last, we obtain $8634$ pairs for training, $900$ for validation, and $7683$ for testing.

\ptitle{Quantitative results.}
As shown in \cref{table:results-shapnet}, our MIRETR achieves the best results on all four metrics, surpassing the baselines by a large margin. GeoTransformer significantly outperforms CoFiNet, which proves the importance of geometric structure to achieving generality. And our method achieves improvements of more than $10$ pp over GeoTransformer on all registration metrics with PointCLM and ECC. Thanks to the instance-ware geometric transformer, the correspondences extracted by our method provide strong instance information, facilitating the correspondence clustering process. At last, our full model surpasses the previous best ECC-based baseline by around $2$ pp on MR, $11$ pp on MP and $7$ pp on MF, demonstrating the strong generality of \ours{} to unseen objects.


\begin{table}[!t]
\centering
\scriptsize
\setlength{\tabcolsep}{4pt}
\begin{tabular}{l|c|ccc}
\toprule
Model & IR($\%$) & MR($\%$) & MP($\%$) & MF($\%$) \\
\midrule
CoFiNet~\cite{yu2021cofinet} + T-Linkage~\cite{magri2014t} & \multirow{4}{*}{39.53} & 4.59 & 3.24 & 3.79 \\
CoFiNet~\cite{yu2021cofinet} + RansaCov~\cite{magri2016multiple} &  & 7.35 & 5.33 & 6.17 \\
CoFiNet~\cite{yu2021cofinet} + PointCLM~\cite{yuan2022pointclm} &  & 23.40 & 19.66 & 21.36 \\
CoFiNet~\cite{yu2021cofinet} + ECC~\cite{tang2022multi} &  & 41.49 & 26.60 & 32.41 \\   
\midrule
GeoTransformer~\cite{qin2022geometric} + T-Linkage~\cite{magri2014t} & \multirow{4}{*}{\underline{69.65}} & 27.68 & 26.40 & 27.02 \\
GeoTransformer~\cite{qin2022geometric} + RansaCov~\cite{magri2016multiple} &  & 41.86 & 48.16 & 44.78 \\
GeoTransformer~\cite{qin2022geometric} + PointCLM~\cite{yuan2022pointclm} &  & 68.09 & 69.09 & 68.58 \\
GeoTransformer~\cite{qin2022geometric} + ECC~\cite{tang2022multi} &  & 78.03 & 64.08 & 70.37 \\
\midrule
\ours{} (\emph{ours}) + T-Linkage~\cite{magri2014t} & \multirow{5}{*}{\textbf{77.13}} & 29.40 & 28.67 & 29.03 \\
\ours{} (\emph{ours}) + RansaCov~\cite{magri2016multiple} &  & 45.63 & 46.85 & 46.23 \\
\ours{} (\emph{ours}) + PointCLM~\cite{yuan2022pointclm} &  & 85.63 & \underline{85.81} & 85.71 \\
\ours{} (\emph{ours}) + ECC~\cite{tang2022multi} &  & \underline{92.56} & 82.08 & \underline{87.01} \\
\ours{} (\emph{ours}, full pipeline) &  & \textbf{94.95} & \textbf{93.94} & \textbf{94.44} \\ 
\bottomrule
\end{tabular}
\vspace{-5pt}
\caption{Evaluation results on ShapeNet benchmark.}
\label{table:results-shapnet}
\vspace{-10pt}
\end{table}


\begin{table}[!t]
\centering
\scriptsize
\setlength{\tabcolsep}{4pt}
\begin{tabular}{l|c|ccc}
\toprule
Model & IR($\%$) & MR($\%$) & MP($\%$) & MF($\%$) \\
\midrule
(a) full \ours{} & \textbf{45.14} & \textbf{38.51} & \textbf{41.19} & \textbf{39.80} \\
(b) w/o instance-aware point matching & 40.52 & 18.15 & 27.54 & 21.88 \\
(c) w/o instance masking block & 36.78 & 17.07 & 45.37 & 24.80 \\
(d) w/o both & 38.81 & 14.13 & 27.44 & 18.65 \\
(e) w/ global attention & 25.98  & 7.72 & 18.86 & 10.96 \\
\bottomrule
\end{tabular}
\vspace{-5pt}
\caption{Ablation studies on ROBI benchmark.}
\label{table:ablation-robi}
\vspace{-18pt}
\end{table}


\subsection{Ablation Studies}
\label{sec:exp-ablation}

We conduct extensive ablation studies on ROBI benchmark to better understand our design choices in \cref{table:ablation-robi}.
We first ablate the instance-aware point matching and extract point correspondences within only the patches of the superpoint correspondences in \cref{table:ablation-robi}~(b), which is significantly worse than the full model.
The poses estimated by this method is unstable due to clustered correspondences, thus the duplicated candidates cannot be effectively removed by NMS.
Next, we remove the instance masking blocks in \cref{table:ablation-robi}~(c) and the instance-aware point matching gathers the points from all neighbor superpoints. This model achieves a good MP, but a significantly worse MR, which means it can register only few instances (2.9 \vs 13.7 average instances per scene).
We then ablate both modules in \cref{table:ablation-robi}~(d), which further degrades the performance.
At last, we replace the geometric encoding block with the global geometric self-attention~\cite{qin2022geometric} in \cref{table:ablation-robi}~(e), which achieves the worst results. This indicates that the superpoint features can be polluted if global context is encoded.
These results have proven the strong efficacy of our designs.
More ablation studies are presented in the supplementary material.


\section{Conclusion}

We have proposed \ours{}, a coarse-to-fine method to extract instance-aware correspondences for multi-instance registration.
At the coarse level, an instance-aware geometric transformer module jointly learns instance-aware superpoint features and predicts per-instance masks. At the fine level, the superpoint correspondences are extended to instance candidates according to the instance masks, where instance-wise point correspondences are extracted. At last, we devise a simple but effective candidate selection and refinement algorithm to obtain the final registrations, bypassing the need of multi-model fitting methods.
Extensive experiments on three public benchmarks have demonstrated the efficacy of our method.
For future work, we would like to extend \ours{} to multi-modal multi-instance registration to boost more applications.

\ptitle{Acknowledgement.} This work is in part supported by the NSFC (62325211, 62132021, 62102435), the Major Program of Xiangjiang Laboratory (23XJ01009), and the National Key R\&D Program of China (2018AAA0102200).
\clearpage

\appendix

\section{Implementation Details}
\label{supp:implementation-details}

\subsection{Network Architecture}

\ptitle{Backbone.}
We use a KPConv-FPN backbone~\cite{thomas2019kpconv} for feature extraction.
We apply the grid subsampling scheme of~\cite{thomas2019kpconv} to downsample the point clouds and generate superpoints and dense points before feeding them into the network.
The input point clouds are first downsampled with a voxel-grid filter of the size of $2.5\text{cm}$ on Scan2CAD, ShapeNet and $0.15\text{cm}$ on ROBI. 
We adopt a $4$-stage backbone in all benchmarks.
After each stage, the voxel size is doubled to further downsample the point clouds. The first and last (coarsest) levels of downsampled points represent the dense points and superpoints intended for matching.
The detailed network configurations are shown in \cref{table:architecture}.

\ptitle{Instance-Aware Geometric Transformer.}
Given the superpoint features $\hat{\textbf{F}}{}^{\mathcal{P}}$ and $\hat{\textbf{F}}{}^{\mathcal{Q}}$ from the backbone, we first use a linear projection $\textbf{W}_{\text{in}}$ to compress the feature dimension from $1024$ to $256$.
$\mathbf{M}^{\mathcal{Q}}_{(0)}$ is initialized as all zeros.
We adopt $N_t = 3$ instance-aware geometric transformer modules to iteratively extract superpoint features and predict instance masks:
\begin{align}
\hat{\textbf{F}}{}^{\mathcal{P}}_{\text{self},(0)} &= \hat{\textbf{F}}{}^{\mathcal{P}} \textbf{W}_{\text{in}}\\
\hat{\textbf{F}}{}^{\mathcal{Q}}_{\text{self},(0)} &= \hat{\textbf{F}}{}^{\mathcal{Q}} \textbf{W}_{\text{in}}\\
\hat{\textbf{F}}{}^{\mathcal{P}}_{\text{self},(t)} & = \mathrm{GeometricEncoder}(\hat{\mathcal{P}}, \hat{\textbf{F}}{}^{\mathcal{P}}_{\text{cross},(t-1)}), \\
\hat{\textbf{F}}{}^{\mathcal{Q}}_{\text{self},(t)} & = \mathrm{GeometricEncoder}(\hat{\mathcal{Q}}, \hat{\textbf{F}}{}^{\mathcal{Q}}_{\text{cross},(t-1)},\mathbf{M}^\mathcal{Q}_{(t-1)}), \\
\hat{\textbf{F}}{}^{\mathcal{P}}_{\text{cross},(t)} & = \mathrm{CrossAtt}(\hat{\textbf{F}}{}^{\mathcal{P}}_{\text{self},(t)},\hat{\textbf{F}}{}^{\mathcal{Q}}_{\text{self},(t)}), \\
\hat{\textbf{F}}{}^{\mathcal{Q}}_{\text{cross},(t)} & = \mathrm{CrossAtt}(\hat{\textbf{F}}{}^{\mathcal{Q}}_{\text{self},(t)}, \hat{\textbf{F}}{}^{\mathcal{P}}_{\text{cross},(t)})\\
\mathbf{M}^\mathcal{Q}_{(t)} & = \mathrm{InstanceMask}(\hat{\mathcal{Q}}, \hat{\textbf{F}}{}^{\mathcal{Q}}_{\text{cross},(t)},\mathbf{M}^\mathcal{Q}_{(t-1)}).
\end{align}

For the geometric structure embedding, we use $\sigma_d = 0.2\text{m}$ on Scan2CAD, ShapeNet and $\sigma_d = 0.02\text{m}$ on ROBI. We use the $\sigma_a = 15^{\circ}$ on all benchmarks.
In the geodesic embedding, we use $\sigma_{geo} = 0.1\text{m}$ on Scan2CAD, ShapeNet  and $\sigma_{geo} = 0.01\text{m}$ on ROBI.  Given the geodesic distance $G_{i, j}$ between $\hat{\textbf{p}}_i$ and $\hat{\textbf{p}}_j$, the pair-wise 
geodesic distance embedding $\textbf{g}^G_{i, j}$ is computed as:
\begin{equation}
\left\{
\begin{aligned}
\textbf{g}^G_{i, j, 2k} & = \sin(\frac{G_{i, j} / \sigma_{geo}}{10000^{2k / d_t}}) \\
\textbf{g}^G_{i, j, 2k+1} & = \cos(\frac{G_{i, j} / \sigma_{geo}}{10000^{2k / d_t}})
\end{aligned},
\right.
\end{equation}
where $d_t$ is the feature dimension. The geodesic embedding $\textbf{g}_{i, j}$ is computed as:
\vspace{-5pt}
\begin{equation}
\textbf{g}_{i, j} = \textbf{g}^G_{i, j}\textbf{W}^G,
\vspace{-5pt}
\label{eq:geodesic}
\end{equation}
where $\textbf{W}^G \in \mathbb{R}^{d_t \times d_t}$ is the projection matrices for the geodesic embedding.

At last, the final $\hat{\textbf{Z}}{}^{\mathcal{P}}$ and $\hat{\textbf{Z}}{}^{\mathcal{Q}}$ are obtained by adopting another linear projection with $256$ channels.
\begin{align}
\hat{\textbf{Z}}{}^{\mathcal{P}} & = \hat{\textbf{F}}{}^{\mathcal{P}}_{\text{cross},(N_t)} \textbf{W}_{\text{out}}, \\
\hat{\textbf{Z}}{}^{\mathcal{Q}} & = \hat{\textbf{F}}{}^{\mathcal{Q}}_{\text{cross},(N_t)} \textbf{W}_{\text{out}}.
\end{align}

\begin{table}[!t]
\scriptsize
\setlength{\tabcolsep}{5pt}
\centering
\begin{tabular}{c|c|c}
\toprule
Stage & Scan2CAD & ROBI \\
\midrule
\multicolumn{3}{c}{\emph{Backbone}} \\
\midrule
\multirow{2}{*}{1} & KPConv($1 \tight{\rightarrow} 64$) & KPConv($1 \tight{\rightarrow} 64$) \\
 & ResBlock($64 \rightarrow 128$) & ResBlock($64 \rightarrow 128$) \\
\midrule
\multirow{3}{*}{2} & ResBlock($64 \rightarrow 128$, strided) & ResBlock($64 \rightarrow 128$, strided) \\
 & ResBlock($128 \rightarrow 256$) & ResBlock($128 \rightarrow 256$) \\
 & ResBlock($256 \rightarrow 256$) & ResBlock($256 \rightarrow 256$) \\
\midrule
\multirow{3}{*}{3} & ResBlock($256 \rightarrow 256$, strided) & ResBlock($256 \rightarrow 256$, strided) \\
 & ResBlock($256 \rightarrow 512$) & ResBlock($256 \rightarrow 512$) \\
 & ResBlock($512 \rightarrow 512$) & ResBlock($512 \rightarrow 512$) \\
\midrule
\multirow{3}{*}{4} & ResBlock($512 \rightarrow 512$, strided) & ResBlock($512 \rightarrow 512$, strided) \\
 & ResBlock($512 \rightarrow 1024$) & ResBlock($512 \rightarrow 1024$) \\
 & ResBlock($1024 \rightarrow 1024$) & ResBlock($1024 \rightarrow 1024$) \\
\midrule
\multirow{2}{*}{5} & NearestUpsampling & NearestUpsampling \\
 & UnaryConv($1536 \rightarrow 512$) & UnaryConv($1536 \rightarrow 512$) \\
\midrule
\multirow{2}{*}{6} & NearestUpsampling & NearestUpsampling \\
 & UnaryConv($768 \rightarrow 256$) & UnaryConv($768 \rightarrow 256$) \\
\midrule
\multicolumn{3}{c}{\emph{Instance-Aware Geometric Attention}} \\
\midrule
1 & Linear($1024 \rightarrow 256$) & Linear($1024 \rightarrow 256$) \\
\midrule
\multirow{2}{*}{2} & Geometric encoding block(256, 4) & Geometric encoding block(256, 4) \\
 & Cross-attention block(256, 4) & Cross-attention block(256, 4) \\
 & Instance masking block(256, 4) & Instance masking block(256, 4) \\
\midrule
\multirow{2}{*}{3} & Geometric encoding block(256, 4) & Geometric encoding block(256, 4) \\
 & Cross-attention block(256, 4) & Cross-attention block(256, 4) \\
 & Instance masking block(256, 4) & Instance masking block(256, 4) \\
\midrule
\multirow{2}{*}{4} & Geometric encoding block(256, 4) & Geometric encoding block(256, 4) \\
 & Cross-attention block(256, 4) & Cross-attention block(256, 4) \\
 & Instance masking block(256, 4) & Instance masking block(256, 4) \\
\midrule
5 & Linear($256 \rightarrow 256$) & Linear($256 \rightarrow 256$) \\
\bottomrule
\end{tabular}
\caption{
Network architecture for Scan2CAD and ROBI.
}
\label{table:architecture}
\end{table}

\subsection{Loss Functions}

Our model is trained with three loss functions, an overlap-aware circle loss $\mathcal{L}_{\text{circle}}$, a negative log-likelihood loss $\mathcal{L}_{\text{nll}}$, and a mask prediction loss $\mathcal{L}_{\text{mask}}$. The overall loss is computed as: $\mathcal{L} = \mathcal{L}_{\text{circle}} + \mathcal{L}_{\text{nll}} + \mathcal{L}_{\text{mask}}$. 

\ptitle{Overlap-aware Circle Loss.}
To supervise the superpoint feature features, we follow~\cite{qin2022geometric} and use the overlap-aware circle loss which weights the loss of each superpoint (patch) match according to their overlap ratio. Given the set of anchor patches $\mathcal{A}$, it consists of the patches in $\mathcal{Q}$ which have at least one positive patch in $\mathcal{P}$. For each anchor patch $\mathcal{G}^{\mathcal{Q}}_i \in \mathcal{A}$, we denote the set of its positive patches in $\mathcal{P}$ which share at least $10\%$ overlap with $\mathcal{G}^{\mathcal{Q}}_i$ as $\varepsilon^i_p$, and the set of its negative patches which do not overlap with $\mathcal{G}^{\mathcal{Q}}_i$ as $\varepsilon^i_n$.
The overlap-aware circle loss on $\mathcal{Q}$ is then computed as:
\begin{equation}
\label{eq:overlap-aware-circle-loss}
\mathcal{L}^{\mathcal{Q}}_{\text{circle}} \tight{=} \frac{1}{\lvert\mathcal{A}\rvert} \sum_{\mathclap{\mathcal{G}^{\mathcal{Q}}_i \in \mathcal{A}}} \log[1 + \sum_{\mathclap{\mathcal{G}^{\mathcal{P}}_j \in \varepsilon^i_p}} e^{\lambda^j_i\beta^{i,j}_p (d^j_i - \Delta_p)} \cdot \sum_{\mathclap{\mathcal{G}^{\mathcal{P}}_k \in \varepsilon^i_n}} e^{\beta^{i,k}_n (\Delta_n - d^k_i)}],
\end{equation}
where $d^j_i \hspace{1pt} = \hspace{1pt} \lVert \hat{\textbf{h}}{}^{\mathcal{Q}}_i \hspace{1pt} - \hspace{1pt} \hat{\textbf{h}}{}^{\mathcal{P}}_j \rVert_2$ is the distance in feature space, $\lambda_i^j = (o^j_i)^{\frac{1}{2}}$ and $o^j_i$ is the overlap ratio between $\mathcal{G}^{\mathcal{P}}_i$ and $\mathcal{G}^{\mathcal{Q}}_j$.
The weights $\beta^{i,j}_p \medium{=} \gamma(d^j_i \medium{-} \Delta_p)$ and $\beta^{i,k}_n \medium{=} \gamma(\Delta_n \medium{-} d^k_i)$ are determined individually for each positive and negative example, using the margin hyper-parameters  $\Delta_p \hspace{1pt} {=} \hspace{1pt} 0.1$ and $\Delta_n \hspace{1pt} {=} \hspace{1pt} 1.4$.
The loss $\mathcal{L}^{\mathcal{P}}_{\text{circle}}$ on $\mathcal{P}$ is computed in the same way. And the overall loss is $\mathcal{L}_{\text{circle}} = (\mathcal{L}^{\mathcal{P}}_{\text{circle}} + \mathcal{L}^{\mathcal{Q}}_{\text{circle}}) / 2$.

\ptitle{Negative Log-likelihood Loss.}
Following ~\cite{sarlin2020superglue}, we use a negative log-likelihood loss on the assignment matrix $\bar{\textbf{Z}}_i$ of each ground-truth superpoint correspondence $\hat{\mathcal{C}}^{*}_i$.

For each $\hat{\mathcal{C}}^{}_i$, we calculate the inlier ratio between matched patches using each ground-truth transformation. Subsequently, we select the transformation corresponding to the highest inlier ratio to estimate a set of ground-truth point correspondences ${\mathcal{C}}^{}_i$ with a matching radius $\tau$. 
The point matching loss for $\hat{\mathcal{C}}^{*}_i$ is computed as:
\begin{equation}
\mathcal{L}_{\text{nll}, i} = -\sum_{\mathclap{{(x, y) \in {\mathcal{C}}^{*}_i}}} \log \bar{z}^i_{x, y} - \sum_{x \in \mathcal{I}_i} \log \bar{z}^i_{x, m_i+1} - \sum_{y \in \mathcal{J}_i} \log \bar{z}^i_{n_i+1, y}
\end{equation}
where $\mathcal{I}_i$ and $\mathcal{J}_i$ are the unmatched points in the two matched patches. The final loss is the average of the loss over all sampled superpoint matches: $\mathcal{L}_{\text{nll}} = \frac{1}{N_g} \sum^{N_g}_{i=1} \mathcal{L}_{p, i}$.

\ptitle{Mask Prediction Loss.}
Following~\cite{milletari2016v}, the mask prediction loss consists of the binary cross-entropy (BCE) loss and the dice loss with Laplace smoothing~\cite{milletari2016v}, which is defined as follows:
\begin{equation}
\label{eq:overlap-aware-circle-loss}
\mathcal{L}_{\text{mask},i} \tight{=} \text{BCE}(m_i,m^{gt}_i)+1-2\frac{m_i \cdot m^{gt}_i+1}{\lvert{m_i}\rvert + \lvert{m^{gt}_i}\rvert+1}
\end{equation}
where $m_i$ and $m^{gt}_i$ are the predicted and the ground-truth instance masks, respectively. The final loss is the average loss over all superpoints: $\mathcal{L}_{\text{mask}} = \frac{1}{N_g} \sum^{N_g}_{i=1} \mathcal{L}_{\text{mask}, i}$.

\subsection{Training and Testing Settings}

MIRETR is implemented in Pytorch ~\cite{paszke2019pytorch} with an NVIDIA RTX 3090Ti. 
We train MIRETR using Adam optimizer~\cite{kingma2014adam} for $60/60/40$ epochs , with initial learning rate $10^{-4}$, momentum $0.98$, and weight decay $10^{-6}$ . The learning rate is exponentially decayed by $0.05$ after each epoch.

We use the matching radius of $\tau \tight{=} 5\text{cm}$ for Scan2CAD, ShapeNet and $\tau \tight{=} 0.3\text{cm}$ for ROBI to determine overlapping during the generation of both superpoint-level and point-level ground-truth matches. The number of neighbors is set to $16$ for Scan2CAD, ShapeNet, and $32$ for ROBI. The confidence threshold $\tau$ of the mask score is set to $0.6$. 
We use the same data augmentation as in~\cite{huang2021predator,yu2021cofinet,qin2022geometric}.
During training, We sample $N_g \tight{=} 128$ ground-truth superpoint matches. We generate the ground-truth masks for the instance-aware geometric transformer module and the Instance Candidate Generation module. We calculate the loss between ground-truth masks and the predicted masks in each iteration of the attention module. During testing, We sample $N_c \tight{=} 128$ superpoint matches. During candidate selection and refinement, we set the threshold of similarity score as 0.7 for ROBI and 0.8 for Scan2CAD, ShapeNet. We calculate the number of max inliers within transformations and remove the transformation whose inliers is fewer than $\tau_3\cdot max\_inlier$. We set $\tau_3$ as 0.2 for ROBI and 0.8 for Scan2CAD, ShapeNet. The acceptance radius $\tau_2$ is $5\text{cm}$ for Scan2CAD, ShapeNet and $0.3\text{cm}$ for ROBI. 

\section{Metrics}
\label{supp:metrics}

Following~\cite{tang2022multi,yuan2022pointclm}, we evaluate our method with three registration metrics: (1) \emph{Mean Recall}, (2) \emph{Mean Precision} and (3) \emph{Mean F$_1$ score}. We also report \emph{Inlier Ratio} (IR) and \emph{mean Intersect over Union} (mIoU).

\emph{Mean Recall} (MR) is the ratio of registered instances over all ground-truth instances. For a pair of source point cloud and target point cloud, $\mathtt{recall}$ is computed as:
\begin{equation}
	\mathtt{recall} = \frac{1}{\lvert I^{\text{gt}} \rvert} \sum\nolimits^{\lvert I^{\text{gt}} \rvert}_{s=1} I^{\text{gt}}_s,
\end{equation}
where $I^{\text{gt}}_s \tight{=} \{0,1\}$ represents whether a ground-truth transformation is successfully registered. For non-symmetric instances, the registration is considered successful when the RRE $\leqslant$ $15^{\circ}$, RTE $\leqslant$ 0.1m on Scan2CAD dataset, and RRE $\leqslant$ $15^{\circ}$, RTE $\leqslant$ 0.006m on ROBI dataset. For symmetric instances, the registration is considered successful when the ADD-S $\leqslant$ 0.1 on both datasets. The mean recall (MR) is the average of all $\mathtt{recall}$ in test set. 

\emph{Mean Precision} (MP) is the ratio of registered instances overall predicted instances. For a pair of source point cloud and target point cloud, $\mathtt{precision}$ is computed as:
\begin{equation}
	\mathtt{precision} = \frac{1}{\lvert I^{\text{pred}} \rvert} \sum\nolimits^{\lvert I^{\text{pred}} \rvert}_{s=1} I^{\text{pred}}_s,
\end{equation}
where $I^{\text{pred}}_s \tight{=} \{0,1\}$ represents whether a predicted transformation is successfully registered. The mean precision (MP) is the average of all $\mathtt{precision}$ in test set. 

\emph{Mean F$_1$ score} (MF) is the harmonic mean of MP and MR. The mean $\mathtt{F1}$ Score (MF) is computed as:
\vspace{-2pt}
\begin{equation}
	\mathtt{MF} = \frac{2 \cdot \mathtt{MR} \cdot \mathtt{MP}}{\mathtt{MR}+\mathtt{MP}}
\end{equation}

\emph{Inlier Ratio} (IR) is the ratio of inlier correspondences among putative correspondences. Given point correspondences $\mathcal{C}$, IR is computed as:
\begin{equation}
\mathrm{IR} = \frac{1}{\lvert \mathcal{C} \rvert} \sum_{(\textbf{p}, \textbf{q}) \in \mathcal{C}} \llbracket \lVert \bar{\textbf{T}}_{k}(\textbf{p}) - \textbf{q} \rVert_2 < \tau_1 \rrbracket,
\end{equation}
where $\llbracket \cdot \rrbracket$ is the Iversion bracket, $\textbf{T}_{k}$ is a ground-truth transformation. We set $\tau_1=0.05m$ on Scan2CAD, ShapeNet datasets and $\tau_1=0.005m$ on ROBI dataset.

\emph{Mean Intersect over Union} (mIoU) is the ratio of the intersect of the predicted and the ground-truth instance masks over the union of them, which measures the quality of the predicted masks. Given the predicted and the ground-truth instance masks $\mathbf{M}^{\text{pred}}$ and $\mathbf{M}^{\text{gt}}$, the mIoU of $\mathbf{M}^{\text{pred}}$ is computed as:
\begin{equation}
\mathrm{mIoU} = \frac{\mathrm{Intersect}(\mathbf{M}^{\text{pred}}, \mathbf{M}^{\text{gt}})}{\mathrm{Union}(\mathbf{M}^{\text{pred}}, \mathbf{M}^{\text{gt}})}.
\end{equation}

\section{Additional Experiments}
\label{supp:experiments}

\subsection{Evaluation Results on ModelNet40}
\label{sec:exp-Modelnet40}


\begin{table}[!t]
	\centering
	\scriptsize
	\setlength{\tabcolsep}{5pt}
	\begin{tabular}{l|c|ccc}
		\toprule
		Model & IR(\%) & MR(\%) & MP(\%) & MF(\%) \\
		\midrule
		CofiNet~\cite{yu2021cofinet}+T-Linkage~\cite{choy2019fully}& \multirow{4}{*}{43.01} & 11.38 & 9.12 & 10.12  \\
		CofiNet~\cite{yu2021cofinet}+RansaCov~\cite{choy2019fully} & & 15.13 & 12.89 & 13.92  \\
		CofiNet~\cite{yu2021cofinet}+PointCLM~\cite{yuan2022pointclm} &  & 42.37 & 38.50 & 40.34  \\
		CofiNet~\cite{yu2021cofinet}+ECC~\cite{tang2022multi} &  & 65.31 & 50.24 & 56.79  \\   
		\midrule
		GeoTransformer~\cite{qin2022geometric}+T-Linkage~\cite{magri2014t} & \multirow{4}{*}{58.63} & 33.03 & 31.01 & 31.98  \\
		GeoTransformer~\cite{qin2022geometric}+RansaCov~\cite{magri2016multiple} & & 40.21 & 48.92 & 44.13  \\
		GeoTransformer~\cite{qin2022geometric}+PointCLM~\cite{yuan2022pointclm} &  & 86.14 & 85.37 & 85.75  \\
		GeoTransformer~\cite{qin2022geometric}+ECC~\cite{tang2022multi} &  & 82.49 & 79.26 & 80.84  \\
		\midrule
		\ours{} (\emph{ours}) +T-Linkage~\cite{magri2014t} & \multirow{5}{*}{63.04} & 40.03 & 43.39 & 41.64  \\
		\ours{} (\emph{ours}) +RansaCov~\cite{magri2016multiple} & & 46.03 & 49.58 & 47.73  \\
		\ours{} (\emph{ours}) +PointCLM~\cite{yuan2022pointclm} & & \underline{99.48} & 98.08 & \underline{98.77}   \\
		\ours{} (\emph{ours}) +ECC~\cite{tang2022multi} & & 98.48 & \underline{98.11} & 98.29 \\
		\ours{} (\emph{ours}, full pipeline) & & \textbf{99.95} & \textbf{99.93} & \textbf{99.94}  \\ 
		\bottomrule
	\end{tabular}
	\caption{
		Evaluation results on Modelnet40.
	}
	\label{table:results-modelnet}
\end{table}

\ptitle{Dataset.}
ModelNet40~\cite{wu20153d} consists of 12311 CAD models of man-made objects from 40 different categories. To verify the generalization ability of MIRETR, we train our method and competitors using 5112 point clouds from 20 categories and test on 1266 point clouds from the other 20 categories. For each meshed CAD model, we downsample 4096 points from it to form the source point cloud and generate 4-16 transformations to form the target point cloud.

\ptitle{Results.}
In \cref{table:results-modelnet},our method achieves a score of 99.94 in MF, demonstrating the capability of our model to estimate the poses of previously unseen objects. Additionally, MIRETR surpasses existing methods in MR, MP, and MF. The multi-model fitting methods that take our correspondences as input outperform those using CoFiNet and GeoTransformer, indicating a higher inlier ratio.
\subsection{More Evaluation Results on Scan2CAD}
\label{sec:exp-frame-scan2cad}

\ptitle{Dataset.}
We further evaluate the performance of MIRETR in incomplete scenes by using scene point clouds reconstructed from single-frame RGBD data. Unlike~\cite{yuan2022pointclm}, we do not replace the object instances with the CAD models but use the original incomplete point clouds. We select $6408$ RGBD frames which contain multiple instances for evaluating the performance of the model under occlusion. These frames contain $807$ objects where $564$ are used for training, $80$ for validation, and $163$ for testing.


\begin{table}[!t]
	\centering
	\scriptsize
	\setlength{\tabcolsep}{5pt}
	\begin{tabular}{l|c|ccc}
		\toprule
		Model & IR(\%) & MR(\%) & MP(\%) & MF(\%) \\
		\midrule
		FCGF~\cite{choy2019fully}+T-Linkage~\cite{magri2014t} & \multirow{4}{*}{9.63} & 17.93 & 5.61 & 8.54  \\
		FCGF~\cite{choy2019fully}+RansaCov~\cite{magri2016multiple} & & 21.58 & 9.86 & 13.53 \\
		FCGF~\cite{choy2019fully}+PointCLM~\cite{yuan2022pointclm} & & 36.28 & 18.81 & 24.77  \\
		FCGF~\cite{choy2019fully}+ECC~\cite{tang2022multi} & & 54.77 & 50.34 & 52.46   \\
		\midrule
		CofiNet~\cite{yu2021cofinet}+T-Linkage~\cite{choy2019fully} & \multirow{4}{*}{26.73} & 44.31 & 12.07 & 18.97  \\
		CofiNet~\cite{yu2021cofinet}+RansaCov~\cite{choy2019fully} & & 55.73 & 27.33 & 36.67  \\
		CofiNet~\cite{yu2021cofinet}+PointCLM~\cite{yuan2022pointclm} &  & 44.41 & 50.85 & 47.41  \\
		CofiNet~\cite{yu2021cofinet}+ECC~\cite{tang2022multi} &  & 74.58 & 26.28 & 38.86  \\   
		\midrule
		GeoTransformer~\cite{qin2022geometric}+T-Linkage~\cite{magri2014t} & \multirow{4}{*}{54.05}  & 70.22 & 54.52 & 61.38  \\
		GeoTransformer~\cite{qin2022geometric}+RansaCov~\cite{magri2016multiple} &  & 75.05 & 70.15 & 72.51  \\
		GeoTransformer~\cite{qin2022geometric}+PointCLM~\cite{yuan2022pointclm} &  & 78.95 & 80.80 & 79.86  \\
		GeoTransformer~\cite{qin2022geometric}+ECC~\cite{tang2022multi} &  & 87.88 & 72.37 & 79.37  \\
		\midrule
		\ours{} (\emph{ours}) +T-Linkage~\cite{magri2014t}  & \multirow{5}{*}{57.40} & 74.12 & 47.61 & 57.97  \\
		\ours{} (\emph{ours}) +RansaCov~\cite{magri2016multiple} & & 79.34 & 73.07 & 76.07  \\
		\ours{} (\emph{ours}) +PointCLM~\cite{yuan2022pointclm} & & 81.36 & \underline{84.30} & 82.80   \\
		\ours{} (\emph{ours}) +ECC~\cite{tang2022multi} & & \textbf{92.79} & 75.68 & \underline{83.36} \\
            \ours{} (\emph{ours}, full pipeline) & & \underline{88.06} & \textbf{84.53} & \textbf{86.26}  \\ 
		\bottomrule
	\end{tabular}
	\caption{
		Evaluation results on single-frame Scan2CAD.
	}
	\label{table:results-frame-scancad}
\end{table}

\ptitle{Results.}
As \cref{table:results-frame-scancad} shows, our method achieves the highest mean precision and mean F1 score. Due to the limited number of instances within the Scan2CAD dataset, the methods based on spatial consistency, such as ECC and PointCLM, can effectively filter noise and cluster instances. Therefore, the combination of our method with ECC achieved the highest average recall. However, the multi-model fitting methods integrated with our model achieve superior performance compared to those combined with other point cloud registration methods.

\subsection{Time Evaluation}

we study the time efficiency of MIRETR on ROBI in \cref{table:supp-times}. The model time is the time for correspondence extraction, and the pose time is for transformation estimation.
As shown in \cref{table:supp-times}, while extracting correspondences, our method is slower than CofiNet and GeoTransformer. However, our method achieves faster pose estimation with $3$ times acceleration over PointCLM and $2$ times over ECC, which demonstrates the time efficiency of our method.


\begin{table}[!t]
    \centering
    \scriptsize
    \setlength{\tabcolsep}{6pt}
    \begin{tabular}{l|ccc|c}
    \toprule
    \multirow{2}{*}{Model} & \multicolumn{3}{c|}{Time(s)} & \multirow{2}{*}{MF($\%$)}\\
    & Model & Pose & Total\\
    \midrule
    CofiNet~\cite{yu2021cofinet}+T-Linkage~\cite{magri2014t} & 0.13 & 3.19 & 3.32 & 0.93 \\
    CofiNet~\cite{yu2021cofinet}+RansaCov~\cite{magri2016multiple} & 0.13 & 0.16 & 0.29 & 1.31 \\
    CofiNet~\cite{yu2021cofinet}+PointCLM~\cite{yuan2022pointclm}  & 0.13 & 0.70 & 0.83 & 1.50 \\
    CofiNet~\cite{yu2021cofinet}+ECC~\cite{tang2022multi}  & 0.13 & 0.15 & 0.28 & 4.66 \\
    \midrule
    GeoTransformer~\cite{qin2022geometric}+T-Linkage~\cite{magri2014t}  & 0.09 & 3.13 & 3.22 & 5.73\\
    GeoTransformer~\cite{qin2022geometric}+RansaCov~\cite{magri2016multiple} & 0.09 & 0.17 & 0.26 & 10.81 \\
    GeoTransformer~\cite{qin2022geometric}+PointCLM~\cite{yuan2022pointclm}  & 0.09 & 0.22 & 0.31 & 18.33  \\
    GeoTransformer~\cite{qin2022geometric}+ECC~\cite{tang2022multi}  & 0.09 & 0.17 & 0.26 & 23.20\\
    \midrule
    \ours{} (\emph{ours}) +T-Linkage~\cite{magri2014t} & 0.30 & 3.07 & 3.34 & 11.20\\
    \ours{} (\emph{ours}) +RansaCov~\cite{magri2016multiple} & 0.30 & 0.17 & 0.47 & 18.38\\
    \ours{} (\emph{ours}) +PointCLM~\cite{yuan2022pointclm} & 0.30 & 0.33 & 0.63  &25.48 \\
    \ours{} (\emph{ours}) +ECC~\cite{tang2022multi} & 0.30 & 0.21 & 0.51 & 28.91\\
    \midrule
    \ours{} (\emph{ours, full pipeline})  & 0.30 & 0.10 & 0.40 &  39.80\\ 
    \bottomrule
    \end{tabular}
    \caption{
    Times on ROBI.
The \emph{model time} is the time for correspondence extraction, while the \emph{pose time} is for transformation estimation.
    }
    \label{table:supp-times}
    \end{table}

\subsection{Additional Ablation Studies}
\label{supp:additional-ablation}

\ptitle{Instance-aware geometric transformer.}
\cref{table:supp-instance-aware-geometric-transformer} demonstrates more ablation studies of the instance-aware geometric transformer.

We first study the influence of different embedding methods in the instance-aware geometric transformer module. We compare three methods: (a) the geometric embedding in both the geometric encoding block and the instance masking block, (b) no embedding in the geometric encoding block and the geodesic embedding in the instance masking block, and (c) the geometric embedding in both the geometric encoding block and the geodesic embedding in the instance masking block. \ours{} achieves the best MF, which means the model (a) and (b) extract fewer instances than \ours{}.

We further study the effectiveness of the masking mechanism in the geometric encoding block. Ablating the masking mechanism (d) leads to a significant drop on MP as the superpoint features are polluted by the context outside the instances.

At last, we compare three methods for predicting confidence score of instance masks: (f) feature concatenation and geodesic embedding $[\hspace{3pt}\mathbf{y}_{i_j} ; {\mathbf{y}}_i;\hspace{3pt}\mathbf{g}_{i, j}\hspace{3pt}]$, (g) feature residuals $[\hspace{3pt}\mathbf{y}_{i_j}-{\mathbf{y}}_i\hspace{3pt}]$, and (h) features residuals and geodesic embedding $[\hspace{3pt}\mathbf{y}_{i_j}-{\mathbf{y}}_i;\hspace{3pt}\mathbf{g}_{i, j}\hspace{3pt}]$. It can be observed that the three methods performs comparably.


\begin{table}[!t]
\centering
\scriptsize
\setlength{\tabcolsep}{3pt}
\begin{tabular}{l|ccc|c}
\toprule
Model & MR(\%) & MP(\%) & MF(\%) & mIOU (\%) \\
\midrule
(a) GME \& GME   &36.63  &42.16  & 39.20 & 69.07    \\ 
(b) None \& GDE   &36.41  &42.30 & 39.13& 67.38   \\ 
(c) GME \& GDE (\emph{ours})   & 38.51 & 41.19 & 39.80& 69.26   \\ 
\midrule
(d) GME w/o mask  &38.44  &36.83  & 37.62 & 65.83    \\ 
(e) GME w/ mask (\emph{ours})  & 38.51 & 41.19 & 39.80& 69.26    \\ 
\midrule
(f) $[\hspace{3pt}\mathbf{y}_{i_j} ; {\mathbf{y}}_i;\hspace{3pt}\mathbf{g}_{i, j}\hspace{3pt}]$   &36.11  &41.60  & 38.66 & 69.20    \\ 
(g) $[\hspace{3pt}\mathbf{y}_{i_j} - {\mathbf{y}}_i\hspace{3pt}]$  &37.80  &43.16  & 40.20& 67.13   \\ 
(h) $[\hspace{3pt}\mathbf{y}_{i_j} - {\mathbf{y}}_i\hspace{3pt};\hspace{3pt}\mathbf{g}_{i, j}\hspace{3pt}]$ (\emph{ours}) &38.51 & 41.19 & 39.80 & 69.26\\
\bottomrule
\end{tabular}
\caption{
Ablation studies of the instance-aware geometric transformer on ROBI. \textbf{GME}: geometric embedding. \textbf{GDE}: geodesic embedding. 
}
\label{table:supp-instance-aware-geometric-transformer}
\end{table}

\ptitle{Neighbors.}
We study the influence of different numbers of neighbors in \cref{table:neighbors-robi}. Along with the decreasing number of neighbors, the performance of the model gradually decreases, especially for MP. When the number of neighbors is small, the Instance Candidate Generation module tends to extract more instances but obtain more wrong transformations, resulting in high recall, and low precision.


\begin{table}[!t]
\centering
\scriptsize
\setlength{\tabcolsep}{10pt}
\begin{tabular}{l|c|ccc}
\toprule
\# Neighbors & \# Inst & MR(\%) & MP(\%) & MF(\%) \\
\midrule
4 & 22.69 & 40.25 & 25.39 & 31.14   \\ 
8 & 20.07 & 41.18 & 29.53 & 34.92   \\ 
16 &15.96  & 40.45 & 35.66 & 37.91   \\ 
32 (\emph{ours}) & 13.70 & 38.51 & 41.19 & 39.80   \\ 
48 & 13.01 & 37.59 & 43.83 & 40.47    \\ 
64 &12.60  &37.13  &45.09  & 40.75   \\ 
\bottomrule
\end{tabular}
\caption{
Ablation studies of the number of neighbors on ROBI.
}
\label{table:neighbors-robi}
\end{table}

%

\ptitle{Candidate selection and refinement.}
We first replace the NMS-based filtering in the candidate selection and refinement with random sampling in \cref{supp:nms-robi}, leading to a significant drop on MR. And note that the increase on MP is due to the duplicated registrations.

We conduct the sensitivity analysis on the similarity threshold in NMS in \cref{supp:nms-robi}. $\tau_s$ controls which instance candidates should be merged. When $\tau_s$ goes from $0.9$ to $0.5$, the MP increases ($34.77$ to $46.18$) but the MR drops ($40.20$ to $35.30$), showing that a too small $\tau_s$ could not remove all duplicated instances.


\begin{table}[!t]
\centering
\scriptsize
\setlength{\tabcolsep}{8pt}
\begin{tabular}{l|ccc}
\toprule
Model  & MR(\%) & MP(\%) & MF(\%) \\
\midrule
MIRETR w/ random sampling & 28.94 & 44.71& 37.24   \\ 
MIRETR w/ NMS (\emph{ours}) & 38.51 & 41.19 & 39.80   \\ 
\midrule
NMS 0.9  & 40.20 & 34.77& 37.29   \\ 
NMS 0.8   & 39.35 & 36.02& 38.67   \\ 
NMS 0.7 (\emph{ours}) & 38.51 & 41.19 & 39.80   \\ 
NMS 0.6  & 36.83 & 43.55& 39.91   \\ 
NMS 0.5   & 35.30 & 46.18 & 40.20   \\ 
\bottomrule
\end{tabular}
\caption{
Ablation studies of the candidate selection and refinement module on ROBI.
}
\label{supp:nms-robi}
\end{table}

\subsection{More Qualitative Results}
\label{supp:additional-vis}

\begin{figure*}[t]
  \begin{overpic}[width=1.0\linewidth]{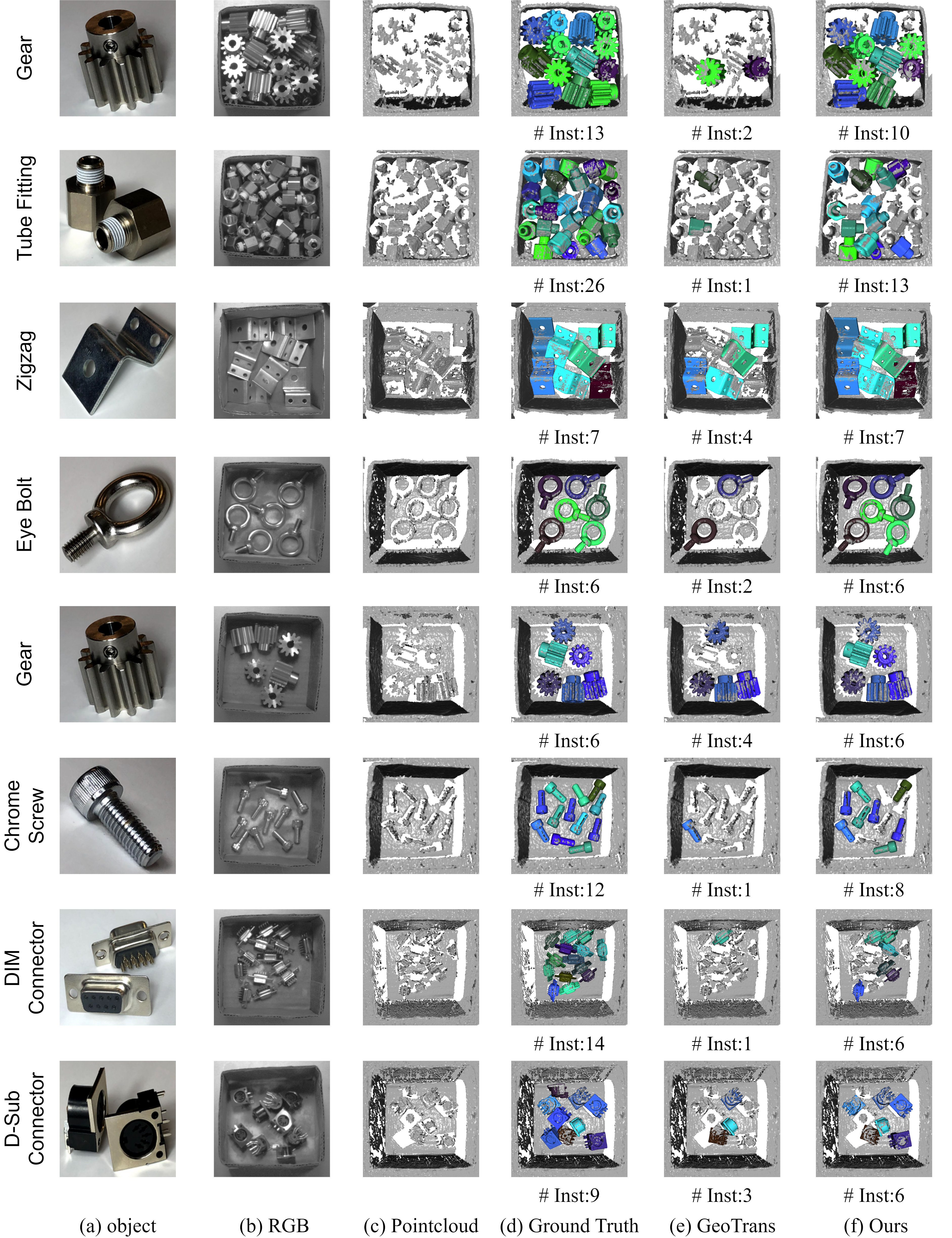}
  \end{overpic}
  \caption{Results on the ROBI dataset. The gray point clouds represent the target point cloud and the point clouds in other colors represent the source point cloud with different transformations. 
  }
  \label{fig:supp_rr1}
\end{figure*}

We provide more qualitative results in \cref{fig:supp_rr1} for ROBI. In \cref{fig:supp_rr1}, our method detects more objects than GeoTransformer~\cite{qin2022geometric}, especially in the low-overlap, clutter scenarios.

\section{Limitations}
In multi-instance scenarios, the amplitude of object rotation changes is greater than in traditional point cloud registration. However, KpConv\cite{thomas2019kpconv} struggles to obtain rotation-invariant features for point matching, which limits our performance. 

In superpoint matching, MIRETR is hampered by the issue of uneven sampling of instances. In some scenarios, MIRETR may sample different points on the same instance. 

\begin{figure*}[t]
  \begin{overpic}[width=1.0\linewidth]{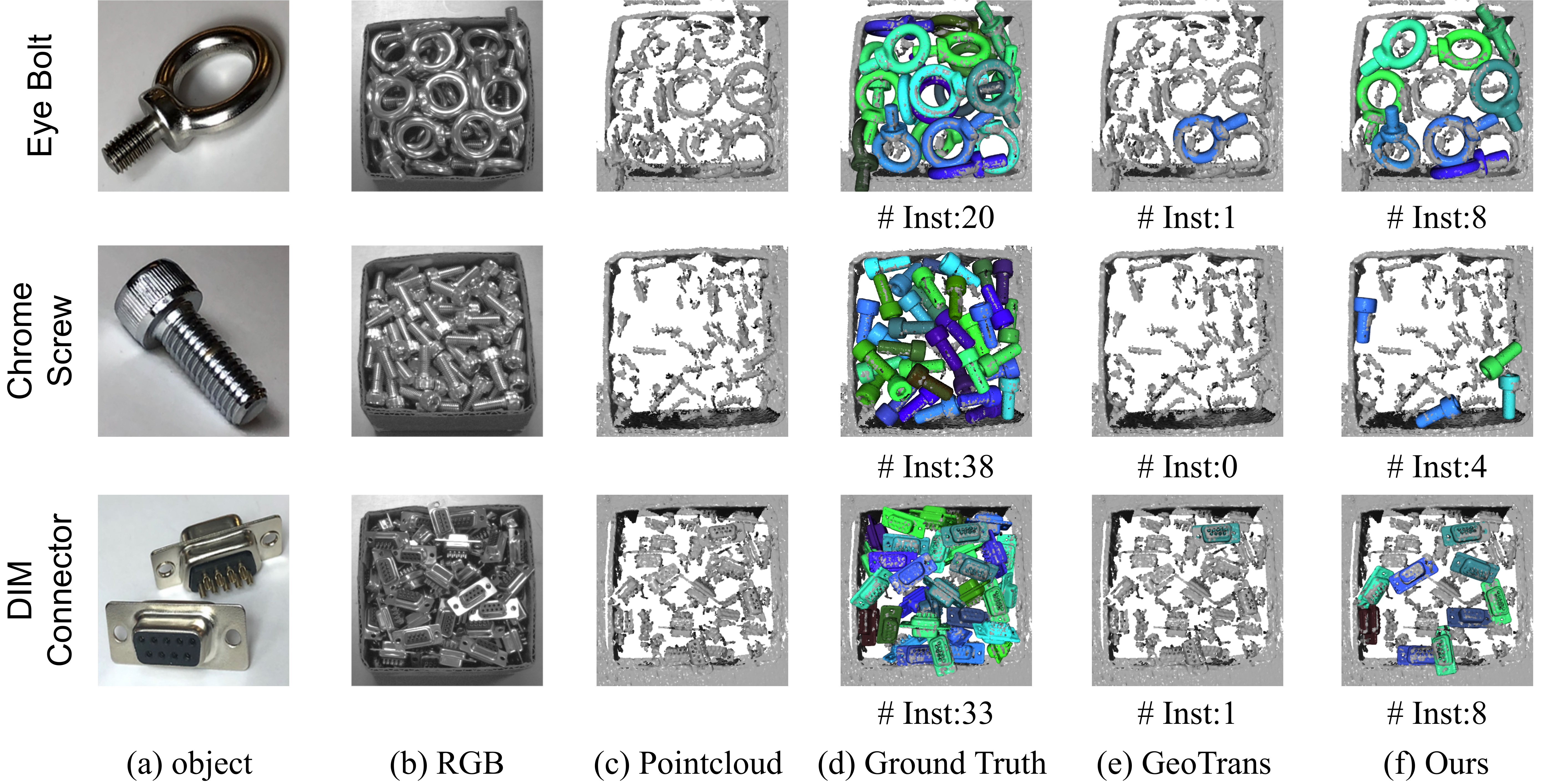}
  \end{overpic}
  \caption{Fail cases of the ROBI dataset. The gray point clouds represent the target point cloud and the point clouds in other colors represent the source point cloud with different transformations. 
  }
  \label{fig:supp_rr}
\end{figure*}

MIRETR is hard to handle extreme clustering and severe occlusion data as shown in~\cref{fig:supp_rr}


{
\small
\bibliographystyle{ieeenat_fullname}
\bibliography{main}
}


\end{document}